\documentclass[a4paper,fleqn]{cas-sc}

\usepackage[numbers,sort&compress]{natbib}

\def\tsc#1{\csdef{#1}{\textsc{\lowercase{#1}}\xspace}}
\tsc{WGM}
\tsc{QE}

\usepackage{amsthm}
\theoremstyle{plain}
\newtheorem{assumption}{Assumption}
\newtheorem{proposition}{Proposition}

\usepackage[caption=false]{subfig}
\usepackage{enumitem}

\begin{document}
\let\WriteBookmarks\relax
\def\floatpagepagefraction{1}
\def\textpagefraction{.001}

\shorttitle{Obstacle-aware spill-free liquid transport}    

\shortauthors{R. Wang et al.}  

\title [mode = title]{Integrated planning and control for obstacle-aware spill-free liquid transport}  



%

\author[1]{Rixin Wang}

\ead{2310114@stu.neu.edu.cn}

\credit{Conceptualization, Methodology, Software, Validation, Formal analysis, Investigation, Visualization, Writing – original draft}

\affiliation[1]{organization={School of Mechanical Engineering and Automation},
            addressline={Northeastern University},
            city={Shenyang},
            postcode={110819},
            state={Liaoning},
            country={China}}

\author[2]{Wenlin Chen}

\ead{chenwenlin@mail.neu.edu.cn}

\credit{Methodology, Software, Validation, Writing – review and editing}

\affiliation[2]{organization={College of Resources and Civil Engineering},
            addressline={Northeastern University},
            city={Shenyang},
            postcode={110819},
            state={Liaoning},
            country={China}}

\author[1]{Shuopeng Wang}

\ead{2210118@stu.neu.edu.cn}

\credit{Investigation, Data curation, Visualization, Writing – review and editing}


\author[1]{Baolin Huang}

\ead{huangbaolin@mails.neu.edu.cn}

\credit{Formal analysis, Validation, Writing – review and editing}

\author[1]{Ying Zhang}

\ead{zhangying@me.neu.edu.cn}

\credit{Resources, Investigation, Writing – review and editing}

\author[1]{Lina Hao}

\cormark[1]

\ead{haolina@me.neu.edu.cn}

\credit{Supervision, Project administration, Funding acquisition, Conceptualization, Writing – review and editing}

\cortext[1]{Corresponding author}



\begin{abstract}
Spill-free liquid transport in cluttered environments requires a robot to satisfy both environmental geometric constraints and spill-free task constraints. Unlike conventional fixed end-effector orientation constraints, the spill-free constraint is time-varying because the liquid response evolves during motion. To address this challenge, this paper presents an integrated planning and control framework. The planning module generates an end-effector reference trajectory according to the slosh-free condition, while the tracking controller applies null-space corrections to follow the reference motion within the robot’s feasible operating range. The two modules are connected by FeasibilityNet, a learning-based initialization module that evaluates, in a single batch, the compatibility between candidate initial joint configurations and the planned reference trajectory and selects the highest-scoring candidate. Simulation results show that the proposed method achieves higher task success rates in cluttered environments than recent related methods while maintaining comparable spill-free performance. Experiments on a seven-degree-of-freedom manipulator with different container–liquid combinations further validate the effectiveness of the proposed framework. 
\end{abstract}


\begin{highlights}
\item Module coordination enables planning and control under multiple coupled constraints.
\item FeasibilityNet ranks initial joint configurations by compatibility with a trajectory.
\item All 30 real-robot trials across six container-liquid combinations are spill-free.
\end{highlights}


\maketitle

\section{Introduction}
Robotic manipulation of deformable materials, particularly liquids, remains challenging \cite{9721534}. Among liquid manipulation tasks, transfer and pouring are widely encountered in practical applications. In life-science and chemistry laboratories, robots are often required to repeatedly move reagents and waste liquids between workstations \cite{10870900,9981636}. In service scenarios, robots are expected to transport drinks safely and smoothly like human waitstaff \cite{biagiotti2023optimalfeedforwardcontrolrobotic, CHOI2024104728}. Although often not the primary goal of a workflow, liquid transport is a fundamental intermediate operation whose quality affects overall system safety, cleanliness, and efficiency. For open containers, spill prevention during transport is essential, as even minor spillage can cause contamination, material loss, or safety risks in laboratory environments and degrade service quality in service applications. Moreover, laboratory and service environments are typically unstructured and cluttered rather than obstacle-free. As a result, practical liquid transport must address spill prevention and obstacle avoidance at the same time. Liquid transport is also a repetitive and labor-intensive operation, and its automation can reduce human workload while improving system efficiency, especially when handling nearly full open containers \cite{jung2026design}. However, for robotic systems, achieving stable and reliable spill-free transport in cluttered environments remains a nontrivial challenge.

First, accurately modeling sloshing dynamics remains challenging. Although fluid-dynamics-based models can, in principle, capture liquid behavior under different container geometries and fluid properties, they are often too computationally expensive for real-time motion planning and control. To improve tractability, many existing methods rely on simplified mechanical surrogates of sloshing \cite{ibrahim2005liquid}, such as pendulum models or mass-spring-damper models. However, these approximations typically require careful parameter identification, and both the model structure and the identified parameters may vary with the container and liquid being used. As a result, such approaches often generalize poorly across task settings.

Second, obstacle-aware spill-free transport requires the simultaneous handling of heterogeneous and strongly coupled constraints. Environmental obstacles impose geometric constraints on robot motion, whereas spill prevention imposes a time-varying orientation constraint on the end effector. Since joint motion affects liquid sloshing, the end-effector orientation must be continuously adjusted during transport to maintain spill-free motion. These two constraints are strongly coupled: collision avoidance may alter joint motion and consequently affect liquid behavior, while spill-free requirements further restrict the feasible collision-free motions of the robot. Nevertheless, most existing studies focus on suppressing liquid oscillation in open environments while neglecting environmental obstacles. Only a few works address spill-free transport and obstacle avoidance simultaneously \cite{8968144,9812387,10802215}. Among them, \citet{8968144} consider obstacle avoidance only for the end-effector trajectory in 3-D task space, while \citet{9812387} reduce collision constraints in a specific scenario to linear inequalities. The sampling-based method in \cite{10802215} enables spill-free transport in cluttered environments, but constrains the path using only a maximum tilt-angle criterion under a quasi-static assumption, without explicitly accounting for the dynamic coupling between robot motion and liquid sloshing. 

Third, achieving task-consistent coordination between motion planning and control remains challenging. Because jointly handling spill-free constraints and whole-body collision avoidance is difficult, existing approaches often adopt a plan-then-track structure \cite{9981173,toth2024rapid,MORIELLO2018129}. This structure is not inherently problematic, but simply connecting a planned trajectory to a generic controller does not ensure successful task execution. The final transport performance depends on both the quality of the generated reference motion and whether the redundant manipulator can safely realize this motion in joint space. A poorly designed reference trajectory may induce excessive sloshing even if it is successfully executed, whereas a favorable reference trajectory may still fail if the corresponding joint-space realization violates kinematic or safety constraints during execution. Planning and control must therefore be designed as complementary components rather than isolated modules.

To address these challenges, we propose an \textbf{i}ntegrated \textbf{p}lanning-and-control framework for \textbf{s}pill-free liquid transport with whole-body \textbf{o}bstacle avoidance (IPSO) in cluttered environments. IPSO addresses the coupled problem using a staged framework consisting of three coordinated modules: 1) Module 1 (Section~\ref{sec:Generation of Slosh-free Reference Trajectories in Cluttered Environments}): end-effector reference trajectory generation, 2) Module 2 (Section~\ref{sec:Learning-Based Evaluation and Selection of Initial Joint Configurations}): initial joint configuration selection, and 3) Module 3 (Section~\ref{sec:Obstacle-Aware Tracking via Null-Space Projection}): task-space trajectory tracking. Modules 1 and 2 are executed offline before transport, while Module 3 operates online to track the preplanned reference trajectory. Instead of relying on a high-fidelity sloshing model, IPSO constrains the end-effector trajectory using a slosh-free condition that is independent of liquid properties, container geometry, and robot-specific parameters \cite{10610813}, thereby enforcing the spill-free objective directly in task space. To handle the coupled spill-free and collision-avoidance requirements, obstacle avoidance is addressed at two complementary levels. Globally, collision-free end-effector reference trajectories guide the manipulator around environmental obstacles. Locally, the manipulator redundancy is exploited to introduce whole-body safety corrections in the null space while preserving the primary spill-free tracking task. The transition from planning to control is further supported by a learning-based initial-configuration selection module, which identifies a joint configuration favorable for the safe execution of the planned reference trajectory. Together, these three modules correspond to three distinct requirements for successful execution: generating a task-consistent reference motion, selecting an initial joint configuration favorable for the joint-space realization, and maintaining safe tracking during online execution. Accordingly, the three-module framework is organized around these execution requirements and their associated failure modes, with the modules coordinated toward the same transport objective.

The main contributions of this paper are summarized as follows:

\begin{enumerate}
\item We formulate obstacle-aware spill-free transport as the safe joint-space realization of a task-space motion subject to coupled slosh-free and obstacle-avoidance constraints and develop a staged framework that coordinates reference-trajectory generation, trajectory-conditioned initialization, and whole-body safety correction.
\item We formulate initial configuration selection for redundant manipulators as a trajectory-conditioned compatibility estimation problem and use a standard multilayer perceptron to estimate the compatibility of candidate joint configurations with reference motions, thereby selecting configurations favorable for downstream execution.
\item We characterize the interaction among the three modules, and derive conditional bounds that quantify how manipulator execution deviations and bounded model deviation affect nominal and actual slosh-free errors.
\item We validate the distinct and complementary roles of three key components of the proposed framework through ablation studies, demonstrate the advantages of the integrated framework through comparisons with external baseline methods, and further conduct real-robot experiments with different container–liquid combinations.
\end{enumerate}

The remainder of this paper is organized as follows. Section~\ref{related works} reviews the related work on spill-free liquid transport using manipulators. Section~\ref{Preliminaries} introduces the slosh-free condition, formulates the problem considered in this paper, and defines the required notation. Section~\ref{method} presents the proposed IPSO framework, detailing the implementation of its three modules and their interactions within the overall framework. Section~\ref{Experiments} reports the simulation and real-world experimental results. Section~\ref{discussion} discusses the limitations of the proposed framework. Finally, Section~\ref{Conclusion} concludes the paper.

\section{Related works}\label{related works}
\subsection{Spill-free liquid transport methods based on sloshing dynamics models}
Liquid sloshing is strongly coupled with robot motion. As a result, many studies on spill-free liquid transport first establish a liquid sloshing model and then perform trajectory planning or controller design based on that model. In robot motion planning and control, a common approach is to approximate sloshing dynamics using equivalent mechanical models, such as pendulum models and mass-spring-damper models \cite{di2022sloshing}. These models describe the liquid response induced by container motion. To simplify analysis and computation, existing studies typically further assume small-amplitude sloshing, under which the free surface of the liquid can be approximated as planar and treated using linear models.

Based on such models, existing methods mainly address slosh-free transport from two perspectives: controller design and trajectory planning. \citet{5650412,https://doi.org/10.1155/2015/279460} use a pendulum model and apply input shaping to the end-effector trajectory as an open-loop feedforward strategy to reduce liquid sloshing. \citet{7989493} design an exponential filter based on a spherical pendulum model to modify the end-effector reference trajectory, which is likewise a feedforward control approach. \citet{REYHANOGLU20131481} develop a full-state feedback controller based on a mass-spring model using Lyapunov stability theory. \citet{8460927} formulate spill prevention as a constraint and proposes an open-loop control method with real-time responsiveness. \citet{9659383} adopt a mass-spring-damper model and reduces liquid oscillation by optimizing the time allocation of the trajectory. More recently, \citet{11692616} consider emergency stopping during liquid manipulation, using a spherical-pendulum sloshing model within a model predictive control (MPC) framework to generate time-optimal stopping trajectories subject to no-spill constraints.

Most of the above methods impose the constraint that the container must remain upright. Although effective for suppressing liquid sloshing, this constraint also significantly restricts the motion flexibility of the manipulator. To allow more flexible container motion, some studies further adopt tilt compensation strategies \cite{8968144,MORIELLO2018129}. These methods actively adjust the container orientation according to the sloshing state so that the liquid remains as stationary as possible relative to the container, thereby enabling slosh-free transport. Among them, \citet{8968144} plan the end-effector motion through optimization and thus follows a trajectory-planning paradigm, whereas \citet{MORIELLO2018129} introduce a filter between planning and control to compute orientation compensation online from a physical model and superimpose it onto the reference trajectory, which is essentially a feedforward control approach.

Overall, these methods rely heavily on sloshing dynamics models. Their effectiveness depends on model accuracy and typically requires parameter identification. Because the model parameters vary with the container, liquid, and operating conditions, these methods generally require model adaptation or parameter re-identification across different tasks \cite{11692616}.

\subsection{Spill-free liquid transport methods with reduced dependence on sloshing dynamics models}

Unlike methods that explicitly rely on sloshing dynamics models, recent studies have explored spill-free trajectory generation without building a complete liquid dynamics model. These methods do not ignore liquid behavior entirely. Instead, they exploit slosh-free motion conditions or simplified couplings between container motion and liquid response, thereby reducing modeling complexity and improving generality.

Inspired by Moroccan tray carrying, \citet{9981173} use a virtual pendulum system to represent slosh-free transport conditions and maps slosh-free trajectory generation to the solution of pendulum motion. Compared with explicit liquid modeling, this approach avoids complex parameter identification and only requires a few geometric parameters of the virtual pendulum. Building on this idea, \citet{toth2024rapid} further introduce an equivalent pendulum model of liquid sloshing into the virtual pendulum system and establishes coupled dynamics between the virtual tray pendulum and the liquid pendulum, enabling more tailored slosh-free trajectory planning for specific container shapes. Similarly, \citet{10610813} model the manipulator end-effector as a virtual quadrotor and exploits differential flatness to generate end-effector orientation trajectories that satisfy slosh-free conditions. This method does not depend on container- or liquid-specific parameters, but instead constructs feasible reference motions in a geometric manner. As a result, it avoids the singularity issues of orientation representations in pendulum models and offers better generality across objects. However, because it does not explicitly model sloshing modes, its suppression performance on some objects may be weaker than methods designed for particular dominant modes. In addition, \citet{10802215} do not explicitly model sloshing dynamics, but instead implicitly captures the relation between container trajectories and spillage risk through learning, and uses it for spill-free trajectory generation in cluttered environments.

Overall, these methods do not aim to construct sloshing dynamics accurately. Instead, they generate spill-free motions through simplified couplings or learned models. Compared with strongly model-dependent approaches, they usually require less modeling effort and offer better generality, making them more practical when the container, liquid, or task setting varies.

In summary, existing spill-free liquid transport methods can be broadly grouped into two categories. One relies on explicit sloshing dynamics models, and the other generates spill-free trajectories through simplified couplings or data-driven learning. Given that container shapes, liquid properties, and environmental conditions often vary significantly in practice, this work builds on a spill-free trajectory generation method with reduced dependence on sloshing dynamics models \cite{10610813}, and further integrates obstacle avoidance and trajectory tracking control into a complete framework for cluttered environments.

\section{Preliminaries and problem formulation}\label{Preliminaries}
\subsection{Slosh-free condition}\label{sec:Physical Basis of Slosh-Free Motion}

This section introduces the apparent-acceleration-based slosh-free condition adopted as a task constraint in IPSO. Related conditions have been used in previous robotic liquid-transport studies \cite{MORIELLO2018129,9981173,10610813}. By coupling the translational acceleration of the container with its orientation, this condition is used to construct a nominal slosh-free reference motion without requiring an explicit container-specific liquid-dynamics model. 

The condition is derived from a simplified pendulum-based
model of liquid sloshing that retains the dominant first
asymmetric sloshing mode and assumes an approximately planar
liquid free surface~\cite{MORIELLO2018129}. These assumptions define a qualitative validity regime: the approximation may lose accuracy for large-amplitude free-surface motion or near-resonant excitation, where nonlinear or nonplanar sloshing responses may arise~\cite{ibrahim2005liquid}. However, these validity conditions cannot generally be converted into universal acceleration or jerk limits, because the liquid response depends not only on the excitation magnitude but also on its frequency content and container--liquid-specific modal properties, such as the natural frequencies and damping~\cite{ibrahim2005liquid,MORIELLO2018129}.

In the following, quantities predicted by this simplified model are referred to as nominal quantities, distinguishing them from the actual liquid response. In this work, we assume that this simplified model provides an adequate description over the evaluated range and focus on motion planning and control under the resulting constraint. Determining a numerical validity boundary for a specific container--liquid system would require system-specific dynamic modeling or direct measurement of the liquid response and is beyond the scope of this work.

Let $a_x$ and $a_z$ denote the horizontal and vertical components of the container linear acceleration $\boldsymbol{a}_e$, respectively. Let $\boldsymbol{g}$ denote the gravitational acceleration vector with magnitude $g=\|\boldsymbol{g}\|$, and let $\theta_{\text{qs}}$ denote the angle between the upward vertical direction and the normal of the liquid free surface. Their relation according to the simplified model is
\begin{equation}
\label{eq:relation_between_sfangle_a}
    (a_z+g)\tan\theta_{\text{qs}}=a_x .
\end{equation}
The apparent acceleration of the moving container is
defined as $\boldsymbol{a}_{\mathrm{app}} = \boldsymbol{a}_e-\boldsymbol{g}$. Accordingly, Eq.~\eqref{eq:relation_between_sfangle_a} indicates that the upward normal of the liquid surface is collinear with
$\boldsymbol{a}_{\mathrm{app}}$, and can be written as
\begin{equation}
    \boldsymbol{n}_{\mathrm{qs}}
    =
    \frac{\boldsymbol{a}_{\mathrm{app}}}
    {\|\boldsymbol{a}_{\mathrm{app}}\|}
    =
    \frac{\boldsymbol{a}_e-\boldsymbol{g}}
    {\|\boldsymbol{a}_e-\boldsymbol{g}\|}.
\end{equation}
Let $\boldsymbol{\eta}\in\mathbb{R}^3$ denote the unit vector along the longitudinal axis of the container, directed toward the container opening. The nominal slosh-free condition is defined as
\begin{equation}
    \boldsymbol{\eta}= \boldsymbol{n}_{\mathrm{qs}} .
\end{equation}
Thus, the end-effector orientation is adjusted according to the container acceleration so that the container axis remains aligned with the free-surface normal. This acceleration--orientation coupling is used in Section~\ref{sec:Generation of Slosh-free Reference Trajectories in Cluttered Environments} to construct the reference motion for the end effector.

Based on this condition, we define the \emph{nominal slosh-free error angle} as
\begin{equation}
e_{\mathrm{sf}}=\arccos\left(\frac{\boldsymbol{\eta}^{\top}\boldsymbol{n}_{\mathrm{qs}}}{\|\boldsymbol{\eta}\|\|\boldsymbol{n}_{\mathrm{qs}}\|}\right).
\end{equation}
This quantity measures how closely the executed container motion satisfies the adopted nominal slosh-free condition. Let $\boldsymbol{n}_{\ell}$ denote the actual upward unit normal of the liquid free surface. We define the \emph{actual slosh-free error angle} as
\begin{equation}
e_{\mathrm{liq}}
=
\arccos\left(
\frac{
\boldsymbol{\eta}^{\top}\boldsymbol{n}_{\ell}
}{
\|\boldsymbol{\eta}\|
\|\boldsymbol{n}_{\ell}\|
}
\right),
\end{equation}
and the model-deviation angle, which characterizes the deviation between the nominal slosh-free model and the actual liquid response, as
\begin{equation}
e_{\mathrm{qs}}
=
\arccos\left(
\frac{
\boldsymbol{n}_{\mathrm{qs}}^{\top}\boldsymbol{n}_{\ell}
}{
\|\boldsymbol{n}_{\mathrm{qs}}\|
\|\boldsymbol{n}_{\ell}\|
}
\right).
\end{equation}
Since $e_{\mathrm{liq}}$, $e_{\mathrm{sf}}$, and $e_{\mathrm{qs}}$ are the pairwise geodesic angular distances among the three directions $\boldsymbol{\eta}$,
$\boldsymbol{n}_{\mathrm{qs}}$, and $\boldsymbol{n}_{\ell}$, the triangle and reverse triangle inequalities on $\mathbb{S}^{2}$ give
\begin{subequations}
\label{eq:liquid_error_bounds}
\begin{align}
\left|e_{\mathrm{liq}}-e_{\mathrm{sf}}\right|
&\leq e_{\mathrm{qs}},
\label{eq:liquid_error_bounds_a}\\
e_{\mathrm{liq}}
&\leq e_{\mathrm{sf}}+e_{\mathrm{qs}}.
\label{eq:liquid_error_bounds_b}
\end{align}
\end{subequations}

In this work, we assume that $e_{\mathrm{qs}}$ remains bounded and sufficiently small over the considered operating regime, such that $e_{\mathrm{sf}}$ provides a close approximation of $e_{\mathrm{liq}}$. Accordingly, $e_{\mathrm{sf}}$ is used in this work to quantify adherence to the nominal slosh-free task constraint and as an indirect measure of slosh-free performance. Section~\ref{sec: Module Interactions} further derives conditional bounds showing how manipulator execution errors and bounded model deviation jointly affect $e_{\mathrm{liq}}$.

\subsection{Notation and problem formulation}\label{sec:Notation and Problem Formulation}
For corresponding actual, reference, and error quantities, unsubscripted symbols denote actual values, while the subscripts \(r\) and \(e\) denote reference values and errors, respectively. Unless otherwise specified, \(\|\cdot\|\) denotes the Euclidean norm for vectors and the induced \(2\)-norm for matrices. The main variables are defined as follows.
\begin{itemize}
\item \(\boldsymbol q(t)\in\mathcal Q\subset\mathbb R^{n_q}\): the manipulator joint configuration at time \(t\). \(\mathcal Q\) denotes the configuration space. The Sawyer robot used in this work has \({n_q}=7\).
\item \(\Sigma_C\): the tool frame attached to the manipulator end-effector, which also serves as the container frame. Its origin is located at the midpoint between the two gripper fingers, and its \(\boldsymbol z\)-axis is normal to the plane defined by the fingers. The \(\boldsymbol z\)-axis of \(\Sigma_C\) is aligned with the longitudinal axis of the container and corresponds to the container axis \(\boldsymbol\eta\) defined in Section~\ref{sec:Physical Basis of Slosh-Free Motion}. All trajectories generated and tracked in the remainder of this paper refer to the motion of \(\Sigma_C\). In this work, we consider grasping-based liquid transport, where the container is rigidly attached to the manipulator end-effector throughout the motion.
\item \(\boldsymbol T(\boldsymbol q)\in\mathrm{SE}(3)\): the pose of \(\Sigma_C\), given by
\[
\boldsymbol T(\boldsymbol q)
=
\begin{bmatrix}
\boldsymbol R(\boldsymbol q) & \boldsymbol p(\boldsymbol q)\\
\boldsymbol 0_{1\times 3} & 1
\end{bmatrix},
\]
where \(\boldsymbol p(\boldsymbol q)\in\mathbb R^3\) and \(\boldsymbol R(\boldsymbol q)\in\mathrm{SO}(3)\) denote the end-effector position and orientation, respectively. Along the joint trajectory \(\boldsymbol q(t)\), we use the shorthand
\[
\boldsymbol p(t)
=
\boldsymbol p(\boldsymbol q(t)),
\quad
\boldsymbol R(t)
=
\boldsymbol R(\boldsymbol q(t)),
\quad
\boldsymbol T(t)
=\boldsymbol T(\boldsymbol q(t)).
\]
The corresponding quantities
\(\boldsymbol T_r(t)\), \(\boldsymbol p_r(t)\), and \(\boldsymbol R_r(t)\) denote the reference end-effector motion generated by Module 1 at time \(t\).
\item \(T\in\mathbb R_{>0}\): the duration of the reference trajectory generated by Module 1.
\item \(\boldsymbol z(t), \boldsymbol z_r(t)\in\mathbb S^2\): the actual and reference directions of the \(\boldsymbol z\)-axis of \(\Sigma_C\) at time \(t\), respectively, defined as
\[
\boldsymbol z(t)
=
\boldsymbol R(t)\boldsymbol e_3,
\quad
\boldsymbol z_r(t)
=
\boldsymbol R_r(t)\boldsymbol e_3,
\]
where \(\boldsymbol e_3=[0,0,1]^\top\) and
\(\mathbb S^2=\{\boldsymbol u\in\mathbb R^3\mid\|\boldsymbol u\|=1\}\).

\item \(\mathcal X=\mathbb R^3\times\mathbb S^2\): the task-relevant end-effector state space, consisting of the end-effector position and tool-axis direction.

\item \(\boldsymbol\chi(t), \boldsymbol\chi_r(t)\in\mathcal X\): the actual and reference task-relevant end-effector states at time \(t\), respectively, defined as
\[
\boldsymbol\chi(t)
=
\bigl(
\boldsymbol p(t),
\boldsymbol z(t)
\bigr),
\quad
\boldsymbol\chi_r(t)
=
\bigl(
\boldsymbol p_r(t),
\boldsymbol z_r(t)
\bigr).
\]
\item \(\mathcal O\): the set of environmental obstacles. In this work, the environment is assumed to be known in advance and to remain static throughout the transport task.

\item \(\mathcal Q_{\mathrm{safe}}(\mathcal O)\subset\mathcal Q\): the set of safe joint configurations satisfying the joint-limit, self-collision, and environment-collision constraints.

\item \(\mathcal F_t\subset\mathcal Q\): the joint-space realization set of the reference task state \(\boldsymbol\chi_r(t)\), defined as
\[
\mathcal F_t
=
\left\{
\boldsymbol q\in\mathcal Q
\mid
\boldsymbol\chi(\boldsymbol q)
=
\boldsymbol\chi_r(t)
\right\}.
\]
Thus, \(\mathcal F_t\) contains all joint configurations that realize the reference task state at time \(t\).

\item \(\mathcal L_{\mathrm{safe}}(\boldsymbol\chi_r;\mathcal O)\): the set of safe joint-space trajectories that realize the reference task-space trajectory \(\boldsymbol\chi_r(\cdot)\), defined as
\[
\mathcal L_{\mathrm{safe}}
(\boldsymbol\chi_r;\mathcal O)
=
\left\{
\boldsymbol q(\cdot)
\;\middle|\;
\boldsymbol q(t)
\in
\mathcal F_t
\cap
\mathcal Q_{\mathrm{safe}}(\mathcal O),
\forall t\in[0,T]
\right\}.
\]
A nonempty \(\mathcal L_{\mathrm{safe}}(\boldsymbol\chi_r;\mathcal O)\) indicates that the reference task motion admits at least one safe joint-space realization.
\end{itemize}

\textit{Problem statement:}
Given an \({n_q}\)-DoF redundant manipulator, a start end-effector position \(\boldsymbol p_{\mathrm{start}}\in\mathbb R^3\), a goal position \(\boldsymbol p_{\mathrm{goal}}\in\mathbb R^3\), and an obstacle set \(\mathcal O\), the objective is to find a joint-space trajectory $\boldsymbol q(t):[0,T]\rightarrow \mathcal Q_{\mathrm{safe}}(\mathcal O)$, such that $
\boldsymbol p(\boldsymbol q(0))
=\boldsymbol p_{\mathrm{start}}$ and $
\boldsymbol p(\boldsymbol q(T))
=
\boldsymbol p_{\mathrm{goal}},
$ while the resulting end-effector motion satisfies the slosh-free condition throughout the transport.

Directly solving this problem is difficult because the joint-space trajectory must simultaneously satisfy whole-body geometric collision-avoidance constraints and a time-varying end-effector orientation constraint coupled to the end-effector translational acceleration. IPSO therefore decomposes the coupled problem into three interacting stages. Although the three modules are based on different mathematical principles, they address subproblems of the same objective. The three modules and their interactions are detailed in the next section.

\begin{figure}[pos=t]
\centering
\includegraphics[scale=0.5]{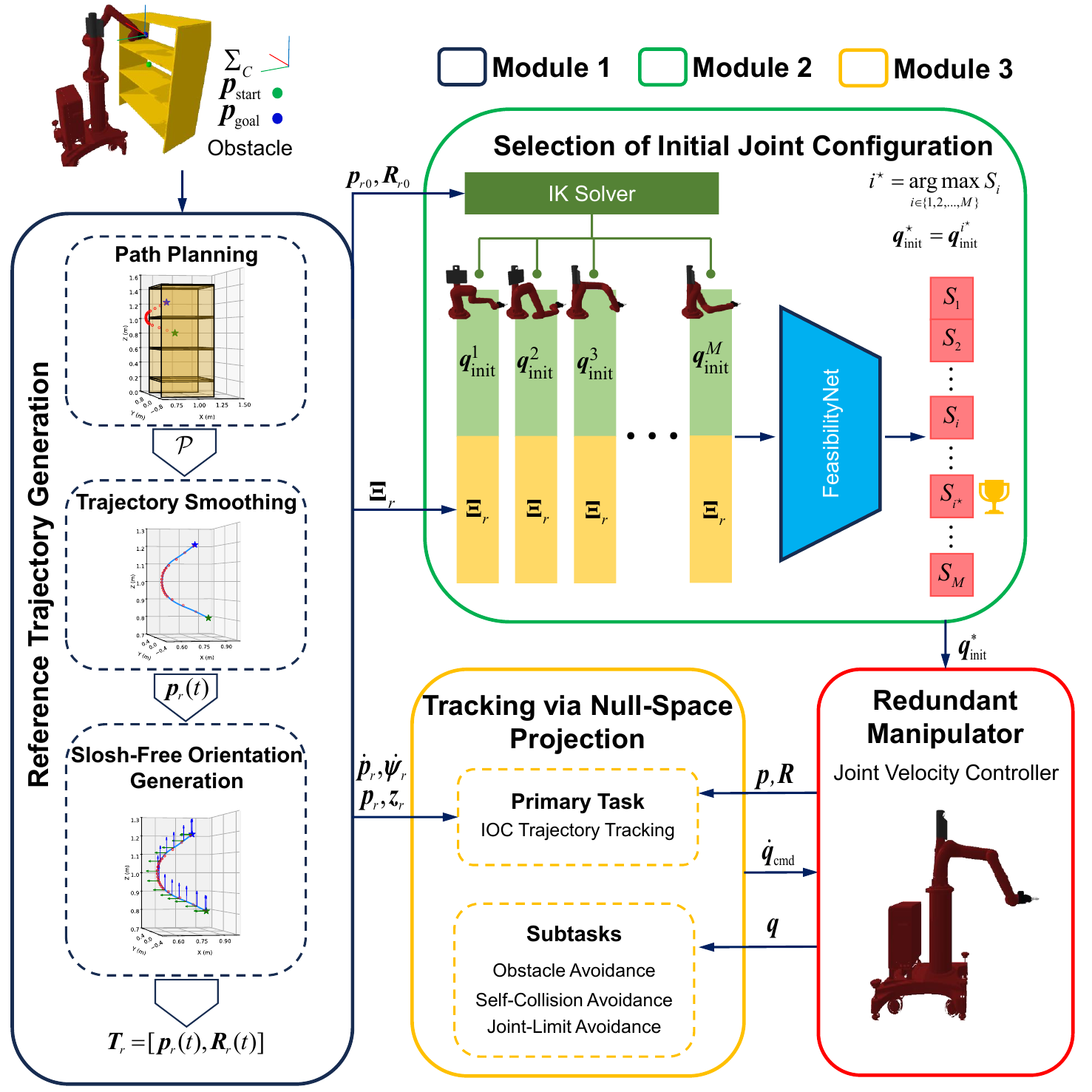} 
\caption{Overview of the proposed framework. Given the start and goal end-effector positions and the environmental obstacles, the framework outputs a manipulator motion that transports the liquid-filled container to the goal position while preventing spillage. Except for $\boldsymbol{p}_{r0}$ and $\boldsymbol{R}_{r0}$, all symbols shown in the figure are defined in the main text. Here, $\boldsymbol{p}_{r0}$ and $\boldsymbol{R}_{r0}$ denote the reference end-effector position and orientation at $t=0$, respectively. The contents of Modules 1–3 are enclosed by blue, green, and yellow rectangles, respectively.}
\label{framework}
\end{figure}

\section{Methodology}\label{method}
An overview of the proposed IPSO framework and the interactions among its three modules is illustrated in Fig.~\ref{framework}.
\subsection{Generation of slosh-free reference trajectories in cluttered environments}\label{sec:Generation of Slosh-free Reference Trajectories in Cluttered Environments}

To generate task-space reference trajectories that satisfy both obstacle-avoidance and slosh-free motion requirements, we follow a three-step procedure: 1) plan a collision-free task-space position path; 2) smooth the path and time-parameterize it to obtain a position trajectory; and 3) generate slosh-free end-effector orientations from the position trajectory and combine them with the position trajectory to form a complete task-space pose trajectory.

\subsubsection{Collision-free task-space path planning}
The objective of this step is to plan a collision-free geometric path in task space connecting \(\boldsymbol{p}_{\mathrm{start}}\) and \( \boldsymbol{p}_{\mathrm{goal}} \): 
\[
\mathcal P
=
\left(
\boldsymbol p_{\mathrm{start}},
\boldsymbol p_1,
\ldots,
\boldsymbol p_{N-2},
\boldsymbol p_{\mathrm{goal}}
\right),
\]
where \(\mathcal{P}\) consists of \(N\) discrete waypoints. This path serves as the geometric basis for generating the reference position trajectory \(\boldsymbol p_r(t)\). To generate the initial collision-free path, we employ a sampling-based planner from the RRT* family. Compared with RRT, RRT* incorporates near-neighbor search and rewiring~\cite{doi:10.1177/0278364911406761}, thereby achieving asymptotic optimality and typically yielding shorter, less tortuous paths that are more amenable to subsequent smoothing.

\subsubsection{B-spline-based trajectory generation and time parameterization}
The goal of this step is to convert the path \(\mathcal P\) into a smooth, time-parameterized, and kinematically feasible reference position trajectory \(\boldsymbol p_r\). B-spline curves are adopted for this purpose because their inherent differentiability and high-order continuity make them well suited for liquid transport tasks.

Given \(n+1\) control points \(\mathbf{P}_0,\mathbf{P}_1,\ldots,\mathbf{P}_n\) and a knot vector \(U=\{u_0,u_1,\ldots,u_m\}\), a trajectory parameterized by a degree-\(p\) B-spline is expressed as a linear combination of \(n+1\) basis functions \cite{shene_bspline_curves}:
\[
\mathbf{C}(u)=\sum_{i=0}^{n}N_{i,p}(u)\mathbf{P}_i
\]
where \(N_{i,p}(u)\in\mathbb{R}\) denotes the degree-\(p\) B-spline basis function, which can be computed recursively using the Cox--de Boor recursion formula. In the task-space setting considered here, \(\mathbf{P}_i\in\mathbb{R}^3\). To ensure that the spline passes exactly through the first and last control points, which correspond to the start and goal positions, and satisfies zero velocity, zero acceleration, zero jerk, and zero snap boundary conditions, we use a clamped B-spline with the knot vector defined as
\[
U = \left[ \underbrace{0, \dots, 0}_{p+1 \text{ times}}, \ \underbrace{\frac{1}{n+1-p}, \dots ,\frac{n-p}{n+1-p}}_{n-p \text{ elements}}, \ \underbrace{1, \dots, 1}_{p+1 \text{ times}} \right],
\]
and set the first five and last five control points as \cite{shene_bspline_derivatives}
\[
\left\{
\begin{aligned}
\mathbf{P}_0 &= \mathbf{P}_1 = \mathbf{P}_2 = \mathbf{P}_3 = \mathbf{P}_4
= \boldsymbol{p}_{\mathrm{start}}\\
\mathbf{P}_n &= \mathbf{P}_{n-1} = \mathbf{P}_{n-2} = \mathbf{P}_{n-3}
= \mathbf{P}_{n-4}
= \boldsymbol{p}_{\mathrm{goal}}
\end{aligned}
\right..
\]

The mapping between the phase variable \(u\in[0,1]\) and time \(t\), denoted by \(u=f^{-1}(t)\), is obtained through time parameterization. We first apply TOPPRA~\cite{8338417} to compute an initial time parameterization of the path. TOPPRA is a time-optimal path parameterization method that accounts for velocity and acceleration constraints. To additionally constrain jerk and snap, we further apply an iterative time-scaling procedure to the initial TOPPRA parameterization. Specifically, a scaling factor \(t_{\mathrm{scale}}>1\) is used to globally stretch the time axis, i.e.,
\[
T_{\text{new}} = t_{\text{scale}} T_{\text{old}}.
\]
During the iteration, \(t_{\text{scale}}\) is gradually increased and the trajectory is recomputed until all kinematic quantities, including velocity, acceleration, jerk, and snap, fall within their upper bounds. This process yields a position trajectory with total duration \(T\).

\subsubsection{Slosh-free orientation generation}

Following~\cite{10610813}, the slosh-free orientation trajectory corresponding to \(\boldsymbol p_r\) is constructed from its second derivative as
\begin{subequations} \label{eq:sf orientation}
\begin{flalign}
& \boldsymbol{a}_{g,r} = \ddot{\boldsymbol{p}}_r(t) - \boldsymbol{g} \\
& \tilde{\boldsymbol{x}} = [\cos\sigma,\sin\sigma,0]^\top \\
& \boldsymbol{z}_r(t)=\frac{\boldsymbol{a}_{g,r}(t)}{\|\boldsymbol{a}_{g,r}(t)\|}\\
&\boldsymbol{y}_r(t)=\frac{\boldsymbol{z}_r(t)\times\tilde{\boldsymbol{x}}}{\|\boldsymbol{z}_r(t)\times\tilde{\boldsymbol{x}}\|}\\
&\boldsymbol{x}_r(t)=\boldsymbol{y}_r(t)\times\boldsymbol{z}_r(t)\\
& \boldsymbol{R}_r(t)=[\boldsymbol{x}_r(t),\boldsymbol{y}_r(t),\boldsymbol{z}_r(t)]
\end{flalign}
\end{subequations}

Here, \(\sigma\) is an auxiliary constant used to construct a complete orthonormal frame, and its value does not affect the slosh-free condition itself. For \(\boldsymbol{y}_r(t)\) to be well defined, it is required that \(\tilde{\boldsymbol{x}} \times \boldsymbol{z}_r(t) \neq \bf{0}\), which can be ensured by choosing \(\sigma\) appropriately.

Through the above procedure, a complete task-space reference pose trajectory \(\boldsymbol T_r\) for \(\Sigma_C\) is obtained. Obstacle avoidance is addressed in the planning of \(\boldsymbol p_r\), while the slosh-free condition is enforced through \(\boldsymbol R_r\).

\subsection{Learning-based evaluation and selection of initial joint configurations}\label{sec:Learning-Based Evaluation and Selection of Initial Joint Configurations}
\begin{figure}[pos=t]
\centering
\includegraphics[scale=1.]{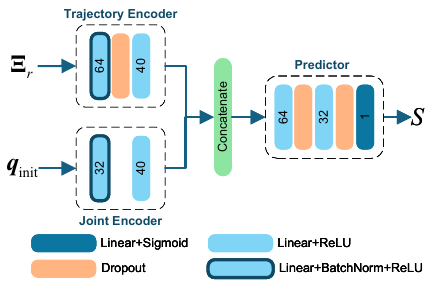} 
\caption{Architecture of FeasibilityNet. The network takes as input the control-point representation
\(\boldsymbol\Xi_r\) of the reference position trajectory, together with a candidate initial joint configuration $\boldsymbol{q}_{\text{init}}$, and outputs a scalar score $S$ that reflects the compatibility of that configuration. The integers inside the rectangles indicate the number of neurons in the corresponding linear layer.}
\label{FNet architecture}
\end{figure}

\subsubsection{Motivation}
Before trajectory tracking, an initial joint configuration
\(\boldsymbol q_{\mathrm{init}}\) must be selected such that
\(\boldsymbol q_{\mathrm{init}}
\in
\mathcal F_0
\cap
\mathcal Q_{\mathrm{safe}}(\mathcal O)\). For a fixed reference trajectory and environment, the resulting closed-loop
joint trajectory depends on the selected initial configuration $\boldsymbol q(t)=\Phi_t\bigl(\boldsymbol q_{\mathrm{init}};
\boldsymbol\chi_r,\mathcal O\bigr),t\in[0,T].$
Although all candidates in \(\mathcal F_0\) realize the same initial task
state, different choices of \(\boldsymbol q_{\mathrm{init}}\) can induce
substantially different joint-space evolutions. An unfavorable choice may
lead the resulting trajectory toward joint limits, self-collision, or
kinematic singularities, thereby degrading tracking performance and
execution success.

Once \(\boldsymbol\chi_r\) has been generated by Module 1, it is treated as
fixed in this stage. Initialization should therefore select the configuration
that is most compatible with the entire reference trajectory. Conceptually,
this can be expressed as
\begin{equation}
\boldsymbol q_{\mathrm{init}}^\star
=
\underset{
\boldsymbol q\in
\mathcal F_0\cap\mathcal Q_{\mathrm{safe}}(\mathcal O)
}{\arg\max}
\;
\mathcal C(\boldsymbol q,\boldsymbol\chi_r),
\label{eq:q_init_star}
\end{equation}
where \(\mathcal C\) denotes the compatibility between an
initial configuration and the reference trajectory. Since this compatibility concerns the entire reference trajectory
\(\boldsymbol\chi_r\), it cannot be reliably characterized by instantaneous metrics
evaluated only at \(t=0\). We therefore train a neural network, termed
FeasibilityNet \(f_\theta(\cdot)\), to evaluate the trackability of each
candidate initial configuration over the entire reference trajectory.

\subsubsection{FeasibilityNet architecture and inference}
To approximate the compatibility function
\(\mathcal C\) in Eq.~\eqref{eq:q_init_star}, FeasibilityNet
\(f_\theta(\cdot)\) maps the control-point representation
\(\boldsymbol\Xi_r\) of the reference position trajectory \(\boldsymbol p_r\) and a candidate
initial configuration \(\boldsymbol q_{\mathrm{init}}\) to a scalar score:
\[
S
=
f_\theta
\bigl(
\boldsymbol q_{\mathrm{init}},\boldsymbol\Xi_r
\bigr)
\approx
\mathcal C
\bigl(
\boldsymbol q_{\mathrm{init}},
\boldsymbol\chi_r
\bigr),
S\in[0,1],
\]
where 
\[
\boldsymbol\Xi_r
=
\begin{bmatrix}
\boldsymbol P_0^\top,\boldsymbol P_1^\top,\dots,\boldsymbol P_n^\top
\end{bmatrix}^\top
\in
\mathbb R^{3(n+1)}.
\] 
The compatibility considered here primarily concerns whether tracking the
reference trajectory drives the manipulator toward joint limits
or self-collision. In the considered transport tasks, the reference orientation \(\boldsymbol R_r(t)\) is determined by the
slosh-free condition and varies only within a limited range, thus providing
limited additional discriminative information for ranking different initial configurations. We therefore use $\boldsymbol\Xi_r$ as the
trajectory input to FeasibilityNet.

At deployment time, the inverse-kinematics solver generates a candidate set $\mathcal Q_{\mathrm{cand}}$ consisting of $M$ joint configurations
\[
\mathcal Q_{\mathrm{cand}}
=
\left\{
\boldsymbol q_{\mathrm{init}}^{(i)}
\right\}_{i=1}^{M}
\subset
\mathcal F_0
\cap
\mathcal Q_{\mathrm{safe}}(\mathcal O),
\]
FeasibilityNet then evaluates each candidate, and the initial configuration for tracking is
selected as
\[
\boldsymbol q_{\mathrm{init}}^\star
=
\underset{
\boldsymbol q\in\mathcal Q_{\mathrm{cand}}
}{\arg\max}
\;
f_\theta(\boldsymbol q,\boldsymbol\Xi_r).
\]
The overall procedure is illustrated in the green block of Fig.~\ref{framework}. The trajectory and the joint configuration are encoded by two separate multilayer perceptron (MLP) branches, whose feature vectors are concatenated and passed to a prediction head to produce the final score. The network architecture is shown in Fig.~\ref{FNet architecture}.

\subsubsection{Training data generation}
Training data are generated in simulation. Using Module 1, we generate
\(K\) reference position trajectories
\(\{\boldsymbol p_{r,i}\}_{i=1}^{K}\)
from randomly sampled start--goal pairs in cluttered environments and encode
each trajectory by its control-point representation \(\boldsymbol\Xi_{r,i}\).
For the \(i\)-th trajectory, the inverse-kinematics solver generates
\(M_t\) candidate initial configurations,
\(
\mathcal Q_{\mathrm{train},i}
=
\left\{
\boldsymbol q_{\mathrm{init},i,j}
\right\}_{j=1}^{M_t}
\subseteq
\mathcal F_{0,i}.
\)
Each candidate is then rolled out using the tracking controller introduced
in Section~\ref{sec:Obstacle-Aware Tracking via Null-Space Projection},
yielding a supervision score \(S_{i,j}\).
During rollout generation, the reference orientation is fixed and the
null-space correction is disabled.
Environmental collisions are excluded from scoring, consistent with the
environment-independent compatibility function
\(\mathcal C(\boldsymbol q,\boldsymbol\chi_r)\). The cluttered environments are used only to generate reference trajectories
representative of the intended deployment scenarios.
The resulting dataset is
\[
\mathcal D
=
\left\{
\left(
\boldsymbol\Xi_{r,i},
\boldsymbol q_{\mathrm{init},i,j},
S_{i,j}
\right)
\;\middle|\;
i=1,\ldots,K,\;
j=1,\ldots,M_t
\right\}.
\]

\subsubsection{Labeling}
For rollout \((i,j)\), let \(H^{\mathrm{self}}_{i,j}\) and
\(H^{\mathrm{joint}}_{\mathrm{D},i,j}\) denote the maximum self-collision and joint-limit metrics, respectively, observed during tracking. Their definitions are given in Section~\ref{sec:Obstacle-Aware Tracking via Null-Space Projection}, specifically in Eqs.~\eqref{eq:H_self} and \eqref{eq:H_D_joint}. The normalized rollout risk is defined as
\[
\mathcal R_{i,j}
=
\min
\left\{
1,
\max
\left(
\frac{H^{\mathrm{self}}_{i,j}}{\eta_{\mathrm{self}}},
\frac{H^{\mathrm{joint}}_{\mathrm{D},i,j}}{\eta_{\mathrm{joint}}}
\right)
\right\},
\]
where \(\eta_{\mathrm{self}}\) and \(\eta_{\mathrm{joint}}\) are the corresponding risk thresholds. The inner maximum implements a worst-case aggregation of the two constraints.

A rollout is labeled as successful if tracking reaches the end of the reference trajectory without self-collision or joint-limit violation; otherwise, it is labeled as failed. The supervision score is defined as
\begin{equation}
S_{i,j}
=
\begin{cases}
\displaystyle
\alpha
\frac{t_{\mathrm{act},i,j}}{T_i},
&
\text{Failure},
\\[6pt]
\displaystyle
\beta
+
(1-\beta)
\left(
1-\mathcal R_{i,j}
\right),
&
\text{Success},
\end{cases}
\end{equation}
where \(t_{\mathrm{act},i,j}\) is the actual execution time and
\(0<\alpha<\beta<1\). Consequently, failed rollouts receive scores in \([0,\alpha)\), whereas successful rollouts receive scores in \([\beta,1)\). Within these two ranges, the score favors later failures and lower-risk successful rollouts, respectively.

\subsection{Obstacle-aware tracking via null-space projection}\label{sec:Obstacle-Aware Tracking via Null-Space Projection}

Task-priority control provides a framework for coordinating primary motion objectives and secondary objectives, such as obstacle avoidance, in redundant manipulators \cite{PETRIC2013948}. Following this principle, we employ a task-space resolved-rate controller with null-space projection to execute the reference trajectory \(\boldsymbol\chi_r\) while incorporating safety-related secondary tasks.
The primary task tracks the reference motion, while null-space corrections
are introduced to improve motion safety during tracking with minimal
interference with the primary task.

\subsubsection{Primary task control}
For the spill-free liquid transport task considered in this work, the
slosh-free condition constrains only the direction of the
\(\boldsymbol z\)-axis of \(\Sigma_C\), while rotation about this axis
remains unconstrained. Therefore, the primary task is to track the
task-relevant reference state
\(\boldsymbol\chi_r(t)=(\boldsymbol p_r(t),\boldsymbol z_r(t))\),
rather than a fully specified end-effector orientation. This corresponds to
an incomplete orientation constraint problem. We therefore adopt the
perpendicular curve-based trajectory tracking controller proposed
in~\cite{9802904}, referred to here as the Incomplete Orientation
Constraint Controller (IOCC).

As described in \cite{9802904}, the commanded joint velocity for the primary task is
\begin{equation}\label{dq_cmd_wo_nullspace}
    \dot{\boldsymbol{q}}_{\text{cmd}}(t)
    =
    \boldsymbol{J}_\lambda^{\dagger}
    \dot{\boldsymbol{x}}_d ,
\end{equation}
where $\boldsymbol{J}_\lambda^{\dagger}
=
\boldsymbol{J}^\top
\left(\boldsymbol{J}\boldsymbol{J}^\top + \lambda \boldsymbol{I}_6\right)^{-1}$ denotes the damped least-squares (DLS) pseudoinverse of the manipulator Jacobian \(\boldsymbol{J}\). \(\lambda>0\) is a damping factor that improves numerical stability near kinematic singularities.  \(\dot{\boldsymbol{x}}_d\) is the desired end-effector velocity defined as
\begin{subequations}\label{xd_dot}
\begin{align}
   &\dot{\boldsymbol{x}}_d(t)
    =
    k_v\boldsymbol{v}_r(t) 
    +
    k_e\boldsymbol{e}_s(t),\label{eq:xd}\\
&\boldsymbol{v}_r(t) =
\begin{bmatrix}
\dot{\boldsymbol{p}}_r(t) \\
\dot{\boldsymbol{\psi}}_r(t)
\end{bmatrix}, \\
&\boldsymbol{e}_s(t)=
\begin{bmatrix}
\boldsymbol{p}_e(t)\\
\bar{\boldsymbol{\psi}}_e(t)
\end{bmatrix}, \\
&\boldsymbol{p}_e(t) = \boldsymbol{p}_r(t)-\boldsymbol{p}(t),\label{eq:position_error}\\
&\bar{\boldsymbol{\psi}}_e(t)
= \log\!\left(\boldsymbol{R}^\top(t)\boldsymbol{R}_\perp(t)\right).
\end{align}
\end{subequations}
Here, \(k_v>0\) and \(k_e>0\) are controller gains,
\(\boldsymbol v_r\) and \(\boldsymbol e_s\) denote the reference task-space velocity and tracking error, respectively, and \(\dot{\boldsymbol\psi}_r\) is the reference angular-velocity term obtained from the reference motion.
The matrix \(\boldsymbol R_\perp(t)\) is the foot-of-perpendicular (FoP) of \(\boldsymbol R(t)\) on the feasible orientation curve associated with \(\boldsymbol z_r(t)\), and
\(\log(\cdot):\mathrm{SO}(3)\rightarrow\mathbb R^3\) is the logarithmic map. Hence, \(\bar{\boldsymbol\psi}_e\) is the local geodesic orientation error from the current orientation to the set of orientations satisfying the desired tool-axis constraint.

\subsubsection{Safety-related subtasks and performance measures}

To improve motion safety without affecting the primary task, we introduce several safety-related subtasks in the null space. Each subtask is defined through a scalar performance measure \(H\), whose value depends on the joint configuration and reflects the corresponding risk level. The gradient of each performance measure with respect to \(\boldsymbol{q}\) is used as the correction direction in the null space.

\paragraph{\textbf{Obstacle avoidance:}}
The manipulator body is approximated by \(N_s\) collision spheres, and the
environment is represented by a signed distance field (SDF). For the \(i\)-th collision sphere with center
\(\boldsymbol c_i(\boldsymbol q)\) and radius \(r_i\), we define the safety
distance as
\begin{equation}\label{safe_distance_env}
d_i^{\mathrm{safe}}
=
r_i+\delta^{\mathrm{env}},
\end{equation}
where \(\delta^{\mathrm{env}}>0\) is a safety margin. The corresponding
environment-collision performance measure is
\[
h_i^{\mathrm{env}}(\boldsymbol q)
=
\mathrm{softplus}
\left(
d_i^{\mathrm{safe}}
-
SDF\!\left(\boldsymbol c_i(\boldsymbol q)\right)
\right).
\]
The overall measure is
\[
H^{\mathrm{env}}(\boldsymbol q)
=
\sum_{i=1}^{N_s}
h_i^{\mathrm{env}}(\boldsymbol q),
\]
with larger values indicating greater collision risk. The corresponding null-space joint velocity is defined by its negative gradient,
\begin{equation}
\dot{\boldsymbol q}_{\mathrm{null}}^{\mathrm{env}}
=
-\nabla_{\boldsymbol q}
H^{\mathrm{env}}(\boldsymbol q),
\end{equation}
which drives the manipulator toward configurations with lower
environment-collision risk.

\paragraph{\textbf{Self-collision avoidance:}}
Let \(\mathcal{S}_c\) denote the set of sphere pairs used for self-collision checking. For each pair
\((i,j)\in\mathcal S_c\), the signed clearance is defined as
\begin{equation}\label{safe_distance_self}
d_{ij}(\boldsymbol q)
=
\left\|
\boldsymbol c_i(\boldsymbol q)
-
\boldsymbol c_j(\boldsymbol q)
\right\|
-
r_i-r_j-\delta^{\mathrm{self}},
\end{equation}
where \(\delta^{\mathrm{self}}>0\) is a safety margin.
The corresponding self-collision performance measure is
\[
h_{ij}^{\mathrm{self}}(\boldsymbol q)
=
\mathrm{softplus}
\left(
-d_{ij}(\boldsymbol q)
\right).
\]
The overall measure is
\begin{equation}
H^{\mathrm{self}}(\boldsymbol q)
=
\max_{(i,j)\in\mathcal S_c}
h_{ij}^{\mathrm{self}}(\boldsymbol q),
\label{eq:H_self}
\end{equation}
with larger values indicating greater self-collision risk. The associated
null-space joint velocity is
\begin{equation}
\dot{\boldsymbol q}_{\mathrm{null}}^{\mathrm{self}}
=
-
\nabla_{\boldsymbol q}
H^{\mathrm{self}}(\boldsymbol q).
\end{equation}

\paragraph{\textbf{Joint limit avoidance:}}
We adopt the joint-limit performance measure in~\cite{9802904}:
\begin{equation}
\label{eq:H_joint}
H^{\text{joint}}(\boldsymbol q)
=
\frac{1}{n_q}
\sum_{i=1}^{n_q}
\frac{
\left(q_i-q_i^{\mathrm{mid}}\right)^2
}{
\left(q_i^{\max}-q_i^{\mathrm{mid}}\right)^2
},
\end{equation}
where \(q_i^{\max}\) and \(q_i^{\min}\) are the upper and lower limits of joint \(q_i\), respectively, and $q_i^{\mathrm{mid}}=\tfrac{1}{2}(q_i^{\min}+q_i^{\max})$.
Larger values of \(H^{\mathrm{joint}}\) indicate greater joint-limit risk. Its gradient is
\[
\nabla_{\boldsymbol q}H^{\text{joint}}(\boldsymbol q)
=
\frac{2}{n_q}
\left[
\frac{q_1-q_1^{\mathrm{mid}}}
{\left(q_1^{\max}-q_1^{\mathrm{mid}}\right)^2},
\dots,
\frac{q_{n_q}-q_{n_q}^{\mathrm{mid}}}
{\left(q_{n_q}^{\max}-q_{n_q}^{\mathrm{mid}}\right)^2}
\right]^\top,
\]
and the associated null-space joint velocity is
\begin{equation}
\dot{\boldsymbol q}_{\mathrm{null}}^{\mathrm{joint}}
=
-
\nabla_{\boldsymbol q}H^{\mathrm{joint}}(\boldsymbol q).
\end{equation}

During dataset construction in Section~\ref{sec:Learning-Based Evaluation and Selection of Initial Joint Configurations}, we instead use a maximum normalized deviation as the joint limit cost,
\begin{equation}
H^{\text{joint}}_{\text{D}}(\boldsymbol{q})
=
\max_{i = 1,\dots,n_q}
\frac{\left( q_i - q_i^{\mathrm{mid}} \right)^2}
     {\left( q_i^{\max} - q_i^{\mathrm{mid}} \right)^2},
\label{eq:H_D_joint}
\end{equation}
so that severe violation by any single joint is explicitly penalized during supervision. By contrast, online control uses the mean-based differentiable measure in Eq.~\eqref{eq:H_joint} in order to maintain smooth joint-space velocity commands and numerical stability.

\subsubsection{Null-space projection and combined control law}

Incorporating both the primary task and the secondary subtasks, the full commanded joint velocity is computed as
\begin{subequations}\label{eq:rrmc_nullspace}
\begin{align}
&\dot{\boldsymbol{q}}_{\text{cmd}}
=
\boldsymbol{J}_\lambda^{\dagger}\dot{\boldsymbol{x}}_d
+
\alpha_{\text{null}}\,
\boldsymbol{N}_\lambda(\boldsymbol{q})
\sum_{k \in \mathcal{T}_s}
w_k \, \dot{\boldsymbol{q}}_{\text{null}}^{k},
\label{eq:combined_control:a} \\
&\boldsymbol{N}_\lambda(\boldsymbol{q})
=
\boldsymbol{I}_7
-
\boldsymbol{J}^{\dagger}_\lambda\boldsymbol{J},
\label{eq:rrmc_nullspace:b}
\end{align}
\end{subequations}
where \(\alpha_{\text{null}}>0\) scales the null-space velocity, \(\mathcal{T}_s=\{\mathrm{env},\,\mathrm{self},\,\mathrm{joint}\}\) is the set of secondary tasks, \(w_k>0\) are fixed subtask weights, and \(\boldsymbol{N}_\lambda(\boldsymbol{q})\) is the null-space projection matrix.

During FeasibilityNet data collection, the null-space term is disabled by setting \(\alpha_{\text{null}}=0\), so that only the primary task controller is used for task space trajectory tracking. This allows the intrinsic compatibility between an initial joint configuration and a reference trajectory to be evaluated without assistance from null-space correction. In other words, if a trajectory can still be completed safely without null-space compensation, the corresponding initialization is regarded as more favorable.

\subsubsection{Boundedness analysis of the tracking controller}
In the controller in Eq.~(\ref{eq:rrmc_nullspace}), a DLS pseudoinverse is used instead of the Moore--Penrose pseudoinverse to improve numerical robustness near kinematic singularities. However, the DLS-based null-space projector is only an approximation of the Moore--Penrose null-space projector. Therefore, one generally has $\boldsymbol{J}\boldsymbol{J}_\lambda^\dagger\neq\boldsymbol{I}$ and $\boldsymbol{J}(\boldsymbol{I}-\boldsymbol{J}_\lambda^\dagger\boldsymbol{J})\neq\boldsymbol{0}$. This means that the null-space correction may leak into the primary task space and introduce a perturbation into the primary tracking dynamics. In the following, we analyze this effect and show that, under standard boundedness and nonsingularity conditions, the resulting tracking error $\boldsymbol{e}_s$ remains ultimately bounded. For analysis, the finite-duration reference trajectory is extended beyond $T$ by holding its terminal reference state constant, i.e., $\boldsymbol{p}_r(t)=\boldsymbol{p}_r(T)$ and $\boldsymbol R_r(t)=\boldsymbol R_r(T)$ for $t \ge T$. Accordingly, $\boldsymbol{v}_r(t)=0$ after the trajectory ends, while the tracking controller remains active to hold the terminal state.

\begin{assumption}
\label{assump:bounded_ref_velocity}
The reference task-space velocity is bounded along the trajectory. Specifically, there exists a positive constant \(B_v>0\) such that
\begin{equation}
\left\|
\boldsymbol{v}_r(t)
\right\|
\le B_v,
\quad \forall t\ge0,
\end{equation}\end{assumption}
This assumption is satisfied in this work because the reference motion is generated from a smooth B-spline trajectory subject to kinematic bounds, and $\boldsymbol{v}_r(t)=0$ for $t \ge T$.

\begin{assumption}
\label{assump:nonsingular_jacobian}
The manipulator Jacobian used in the controller remains full row rank and bounded. Specifically, there exist positive constants \(\underline{\sigma}>0\) and \(\overline{\sigma}>0\) such that
\begin{equation}
\sigma_{\min}\!\left(\boldsymbol{J}(\boldsymbol{q}(t))\right)
\ge
\underline{\sigma},
\quad
\sigma_{\max}\!\left(\boldsymbol{J}(\boldsymbol{q}(t))\right)
\le
\overline{\sigma},
\quad
\forall t\ge0,
\end{equation}
where \(\sigma_{\min}(\cdot)\) and \(\sigma_{\max}(\cdot)\) denote the minimum and maximum singular values of the Jacobian, respectively.
\end{assumption}

\begin{assumption}
\label{assump:bounded_nullspace}
The weighted null-space correction term is bounded. Specifically, we define
\begin{equation}
\boldsymbol{v}_s(t)
=
\alpha_{\mathrm{null}}
\sum_{k\in\mathcal{T}_s}
w_k \dot{\boldsymbol{q}}_{\mathrm{null}}^k(t).
\end{equation}
There exists a positive constant \(B_{v_s}>0\) such that
\begin{equation}
\|\boldsymbol{v}_s(t)\|
\le
B_{v_s},
\quad
\forall t\ge0,
\end{equation}
\end{assumption}
This assumption is reasonable because the joint velocity limits prevent $\boldsymbol{v}_s(t)$ from becoming unbounded.

\begin{assumption}
\label{assump:local_orientation_error}
The orientation tracking error remains within a compact local region where the logarithmic map is smooth along the trajectory execution. Specifically, there exists a constant $0<\bar{\psi}<\pi$ such that
\begin{equation}
\left\|
\bar{\boldsymbol{\psi}}_e(t)
\right\|
\le
\bar{\psi},
\quad
\forall t\ge0,
\end{equation}
\end{assumption}
This assumption is reasonable for the tracking tasks considered in this work, since trajectory execution starts from an initial configuration consistent with the reference pose and the controller subsequently tracks a continuously varying reference orientation. Therefore, the orientation tracking error is expected to remain within a local region bounded away from the $\pi$-rotation singularity of the logarithmic map throughout the closed-loop execution.

\begin{proposition}
\label{prop:uub_dls_tracking_error}
Suppose that \textup{\textbf{Assumptions~1--4}} hold. Under the sufficient condition specified in the Supplementary Material S1, the tracking error \(\boldsymbol{e}_s(t)\) is uniformly ultimately bounded. More specifically, there exist positive constants \(a_{\lambda}>0\) and \(b_{\lambda}>0\) such that
\begin{equation}
\limsup_{t\to\infty}
\left\|
\boldsymbol{e}_s(t)
\right\|
\le
\frac{b_{\lambda}}{a_{\lambda}} .
\label{eq:main_uub_bound}
\end{equation}\end{proposition}
The explicit expressions of these constants and the detailed proof of \textup{\textbf{Proposition~1}} are provided in Supplementary Material S1.

\subsection{Module interactions and conditions for system success} \label{sec: Module Interactions}
This section analyzes the interactions among the three modules of IPSO and
identifies the conditions under which the overall system is expected to
succeed. The module interactions are analyzed as follows.
\subsubsection{Interaction between Modules 1 and 2}Module 2 conditions the selection of the initial configuration
\(\boldsymbol q_{\mathrm{init}}^\star\) on the reference trajectory
\(\boldsymbol\chi_r\) generated by Module 1, as formulated in
Eq.~\eqref{eq:q_init_star}. In turn, the selected
\(\boldsymbol q_{\mathrm{init}}^\star\) affects the subsequent closed-loop
joint-space evolution \(\boldsymbol q(t)\), and therefore the quality with
which \(\boldsymbol\chi_r\) is executed. Hence, the two modules are coupled
through trajectory-conditioned initialization:
\textbf{\(\boldsymbol\chi_r\) guides the selection of
\(\boldsymbol q_{\mathrm{init}}^\star\), while
\(\boldsymbol q_{\mathrm{init}}^\star\) influences the realization of
\(\boldsymbol\chi_r\)}.

\subsubsection{Interaction between Modules 1 and 3}
Module 1 determines the task-space motion demand to be realized by Module 3.
Let the reference position trajectory be written as
\(\boldsymbol p_r(t)=\boldsymbol \gamma(u(t))\), where
\(\boldsymbol \gamma(u)\) denotes the geometric path and \(u(t)=f^{-1}(t)\) its time
parameterization. Its derivatives satisfy
\[
\ddot{\boldsymbol p}_r
=
\boldsymbol \gamma''(u)\dot u^2
+
\boldsymbol \gamma'(u)\ddot u,
\]
and
\[
\boldsymbol p_r^{(3)}
=
\boldsymbol \gamma'''(u)\dot u^3
+
3\boldsymbol \gamma''(u)\dot u\ddot u
+
\boldsymbol \gamma'(u)\dddot u.
\]
Here,
\(\dot{(\cdot)} = \mathrm{d}(\cdot)/\mathrm{d}t\)
and
\((\cdot)' = \mathrm{d}(\cdot)/\mathrm{d}u\),
while
\(\boldsymbol p_r^{(3)}
=
\mathrm{d}^3\boldsymbol p_r/\mathrm{d}t^3\)
denotes the reference jerk.
Moreover, the rate of change of the slosh-free tool-axis direction satisfies
\[
\left\|\dot{\boldsymbol z}_r\right\|
=
\frac{
\left\|
\left(
\boldsymbol I_3
-
\boldsymbol z_r\boldsymbol z_r^\top
\right)
\boldsymbol p_r^{(3)}
\right\|
}{
\left\|\boldsymbol a_{g,r}\right\|
}.
\]
Consequently, a geometrically more tortuous or more aggressively timed
reference position trajectory can increase
\(\boldsymbol p_r^{(3)}\) and, when the resulting jerk contains a larger
component orthogonal to \(\boldsymbol z_r\), induce faster variation of the
reference orientation. Thus, \textbf{Module 1 determines not only the translational tracking demand of Module 3 but also its rotational tracking demand through the slosh-free
orientation construction}.

\subsubsection{Interaction between Modules 2 and 3}
For each performance measure \(H^k(\boldsymbol q)\), referred to as a safety cost here, larger values indicate a less safe configuration. Its rate of change under the commanded joint velocity can be decomposed as
\begin{equation}
\dot H^k
=
\left(\nabla_{\boldsymbol q}H^k\right)^\top
\boldsymbol J_\lambda^\dagger
\dot{\boldsymbol x}_d
+
\left(\nabla_{\boldsymbol q}H^k\right)^\top
\boldsymbol N_\lambda
\boldsymbol v_s .
\label{eq:H_dot}
\end{equation}
Equation~\eqref{eq:H_dot} reveals the interaction between
Modules 2 and 3. The first term represents the safety-cost variation induced by primary-task
execution. By carefully selecting
\(\boldsymbol q_{\mathrm{init}}\), Module 2 biases the joint-space realization of the reference trajectory toward one with a more favorable primary-task-induced safety-cost evolution over the complete trajectory. Once execution begins, Module 3 acts online through
the second term to locally counteract unsafe tendencies. Thus, \textbf{Module 2 performs trajectory-level branch selection, whereas Module 3 provides local online correction during execution}.

To establish error bounds, we consider a finite motion interval over which the tool-axis tracking error and translational acceleration tracking error remain bounded, while the magnitude of the reference apparent acceleration $\boldsymbol a_{g,r}$ remains bounded away from zero. The condition $\bar a_e<\underline a_g$ imposes a small-perturbation requirement, ensuring that the acceleration tracking error does not cancel $\boldsymbol a_{g,r}$. The deviation
between the nominal slosh-free model and the actual liquid
response is characterized by a bounded model-deviation angle
$e_{\mathrm{qs}}(t)\leq \bar e_{\mathrm{qs}}$, where
$\bar e_{\mathrm{qs}}$ is a nonnegative constant. Under these conditions, the above analysis leads to the following proposition.

\begin{proposition}[Conditional bounds on the nominal and actual slosh-free errors]
\label{prop:slosh_error_bounds}
Let $\delta\ddot{\boldsymbol p}(t)=\ddot{\boldsymbol p}(t)-\ddot{\boldsymbol p}_r(t)$ denote the translational acceleration tracking error, and let $e_z(t)=\arccos\!\left(
\boldsymbol z(t)^{\top}\boldsymbol z_r(t)\right)$
denote the tool-axis tracking error. Suppose that there exist
constants $\bar e_z\geq 0$, $\bar a_e\geq 0$, and $\underline a_g>0$ such that, for all $t$ in the considered motion interval, $e_z(t)\leq \bar e_z,\left\|\delta\ddot{\boldsymbol p}(t)\right\|
\leq \bar a_e, \left\|\boldsymbol a_{g,r}(t)\right\|
\geq \underline a_g$, with $\bar a_e<\underline a_g$.
Then, the nominal slosh-free error angle satisfies
\begin{equation}
e_{\mathrm{sf}}(t)
\leq
\bar e_z
+
\arcsin\!\left(
\frac{\bar a_e}{\underline a_g}
\right).
\label{eq:esf_bound}
\end{equation}

If, in addition, the model-deviation angle satisfies $e_{\mathrm{qs}}(t)\leq\bar e_{\mathrm{qs}}$ for some constant $\bar e_{\mathrm{qs}}\geq 0$, then the actual
slosh-free error angle satisfies
\begin{equation}
e_{\mathrm{liq}}(t)
\leq
\bar e_z
+
\arcsin\!\left(
\frac{\bar a_e}{\underline a_g}
\right)
+
\bar e_{\mathrm{qs}}.
\label{eq:eliq_bound}
\end{equation}
\end{proposition}

The first bound provides a robustness guarantee for the nominal slosh-free error angle $e_{\mathrm{sf}}$ under bounded tool-axis and translational acceleration tracking errors. The second bound extends this result to the actual slosh-free error by
showing that a bounded deviation between the nominal slosh-free model
and the actual liquid response contributes only an additive error
term. Therefore, as long as the model deviation characterized by
$e_{\mathrm{qs}}$ remains bounded, $e_{\mathrm{liq}}$ also remains
bounded with a correspondingly bounded degradation.

The bounds also clarify the interaction among the three modules. Module 1 determines the reference quantities \(\boldsymbol a_{g,r}\) and \(\boldsymbol z_r\), whereas Modules 2 and 3 affect the resulting tracking errors \(e_z\) and \(\delta\ddot{\boldsymbol p}\). Consequently, the three modules jointly influence $e_{\mathrm{sf}}$ and, through $e_{\mathrm{sf}}$, the actual liquid-surface alignment error $e_{\mathrm{liq}}$. The proof of \textbf{Proposition~\ref{prop:slosh_error_bounds}} is provided in Supplementary Material S2.

The overall system is expected to succeed when the following conditions are jointly satisfied.

\noindent\textbf{C1. Reference trajectory feasibility.}
The reference trajectory generated by Module 1 must be both kinematically admissible and safely realizable. Its position component \(\boldsymbol p_r\) must remain within the kinematic tracking capability of the system, with bounded velocity, acceleration, jerk, and snap, and the complete reference trajectory must satisfy $\mathcal L_{\mathrm{safe}}
(\boldsymbol\chi_r;\mathcal O)
\neq\varnothing$.

\noindent\textbf{C2. Favorable initialization.}
Module 2 must select an initial configuration that is sufficiently compatible with the reference trajectory, so that the subsequent joint-space realization remains in a region favorable for safe execution.

\noindent\textbf{C3. Sufficient online correction capability.}
When the robot approaches an unsafe region, the null-space correction must
be sufficient to counteract further unsafe evolution that may lead to task
failure.

\section{Experiments}\label{Experiments}

This section experimentally evaluates IPSO by addressing the following four questions:
\begin{description}
    \item[\textbf{Q1:}] Do the three key components of IPSO perform their intended roles, and do they complement one another in improving overall task execution?
    \item[\textbf{Q2:}] Does the staged planning-and-control framework of IPSO provide an advantage in handling tasks involving multiple coupled constraints?
    \item[\textbf{Q3:}] How does IPSO perform under different levels of motion demand, and what trade-off arises between transport efficiency and nominal slosh-free tracking?
    \item[\textbf{Q4:}] Does IPSO remain effective when deployed on a real robotic system?
\end{description}

\subsection{Implementation details}
\subsubsection{Software and hardware setup}
\begin{figure}[pos=t]
\centering
\includegraphics[scale=0.37]{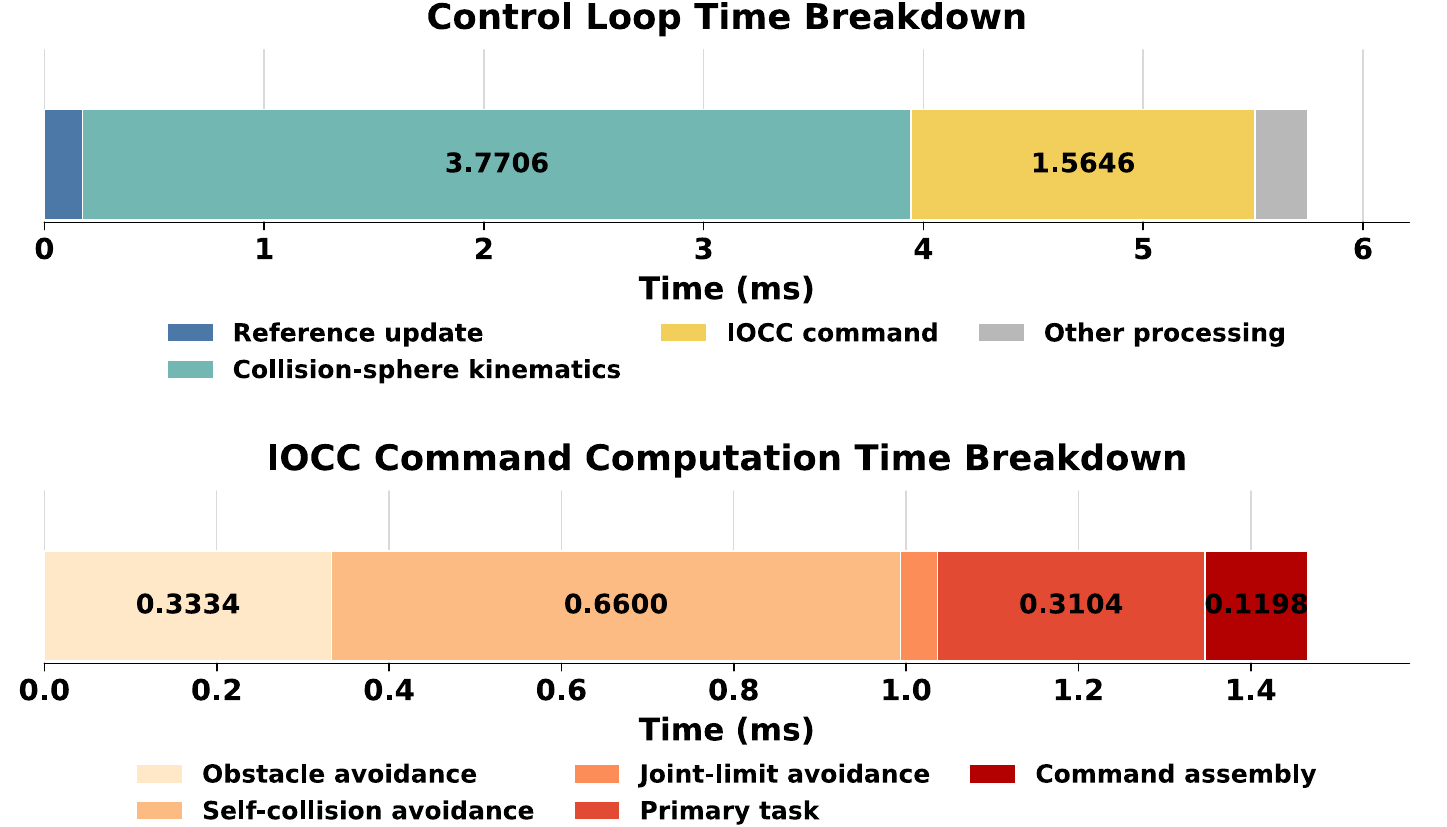} 
\caption{Average per-cycle computation-time breakdown of the control loop measured during a real-robot trajectory. (Top) The complete control-loop computation is divided into reference update, collision-sphere position and Jacobian computation, IOCC command computation, and other processing. (Bottom) The IOCC command computation is further divided into the primary-task command, three null-space corrections for obstacle avoidance, self-collision avoidance, and joint-limit avoidance, and final command assembly. The SDF queries are included in the computation of the obstacle-avoidance null-space correction.}
\label{contro_time_breakdown}
\end{figure}
The entire framework is implemented in Python. For path planning, we use the Informed RRT* algorithm provided by OMPL \cite{6377468}. To compute null-space joint velocities, we adopt the foam tool \cite{2025foamtoolsphericalapproximation} to generate spherical approximations of robot geometry. To balance computational accuracy and efficiency in online tracking control, the manipulator is approximated by 17 collision spheres. In both simulation and real-world experiments, the geometries and poses of environmental obstacles are known in advance. The SDF and its spatial gradients are precomputed before trajectory execution and remain unchanged throughout execution. During control, we leverage the batch-processing capability of TorchRobotics \cite{le2023accelerating,10342382} to query SDF values and their gradients in real time and efficiently compute the spatial positions and Jacobians of these 17 spheres.

The control framework runs at 100 Hz in both simulation and hardware experiments. On the real robot, the control loop typically takes 5–7 ms per cycle. Fig.~\ref{contro_time_breakdown} presents a detailed computation-time breakdown obtained from a representative real-robot trajectory, for which the mean computation time was 5.75 ms per cycle. Collision-sphere kinematics, including the computation of the sphere positions and Jacobians, accounted for the largest portion of the overall control-loop time. Within the IOCC command computation, the self-collision-avoidance correction was the most time-consuming component, whereas the SDF-based obstacle-avoidance correction required approximately half as much computation time on average. Simulations are conducted on a desktop computer equipped with a 13th Gen Intel Core i7-13700KF CPU, 32 GB RAM, and an NVIDIA GeForce RTX 4080 GPU. The simulation environment is built on the open-source simulator Swift \cite{Swift} to simulate the Sawyer manipulator and surrounding obstacles. Real-world experiments are conducted on a 7-DoF Sawyer robot through the Intera SDK.

\subsubsection{Simulation scenarios and evaluation metrics}
\begin{figure}[pos=t]
    \centering
    
    \subfloat[Shelf\label{fig:sub_a}]{
        \includegraphics[width=0.45\columnwidth]{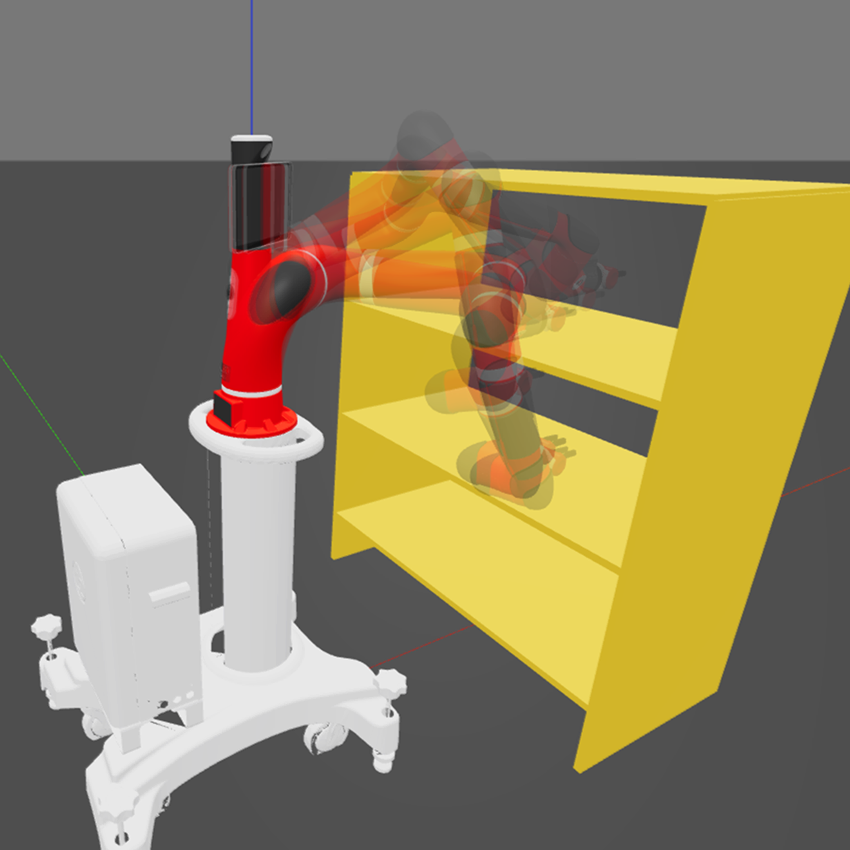}
    }
    \hfill
    \subfloat[Counter\label{fig:sub_b}]{
        \includegraphics[width=0.45\columnwidth]{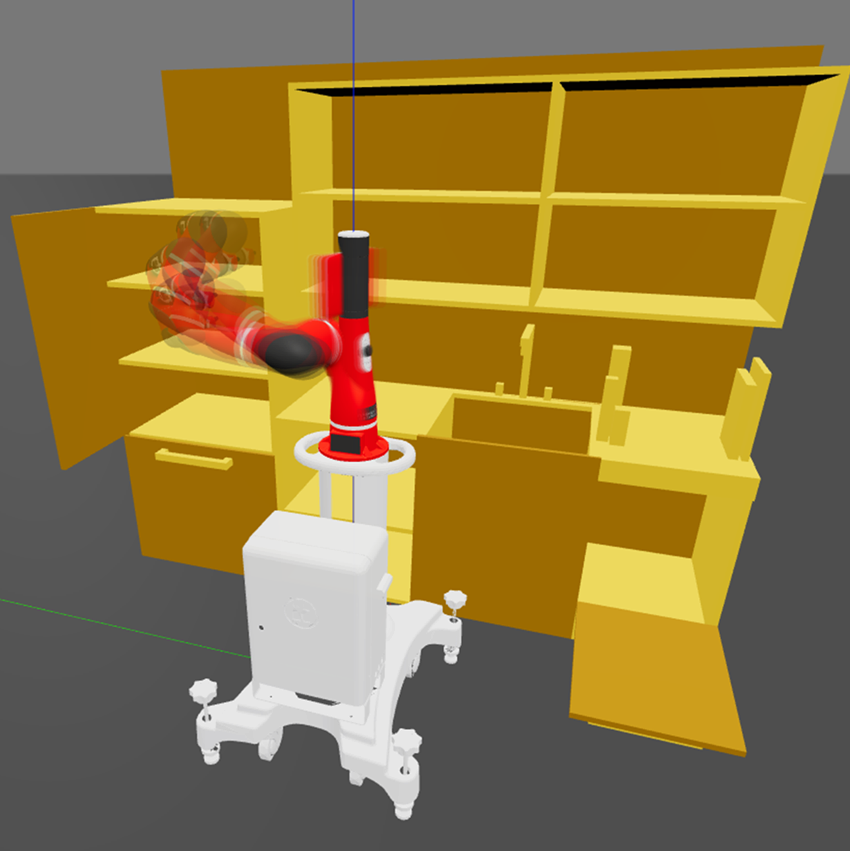}
    }
    
    \vspace{0.5em}
    
    \subfloat[Industrial\label{fig:sub_c}]{
        \includegraphics[width=0.45\columnwidth]{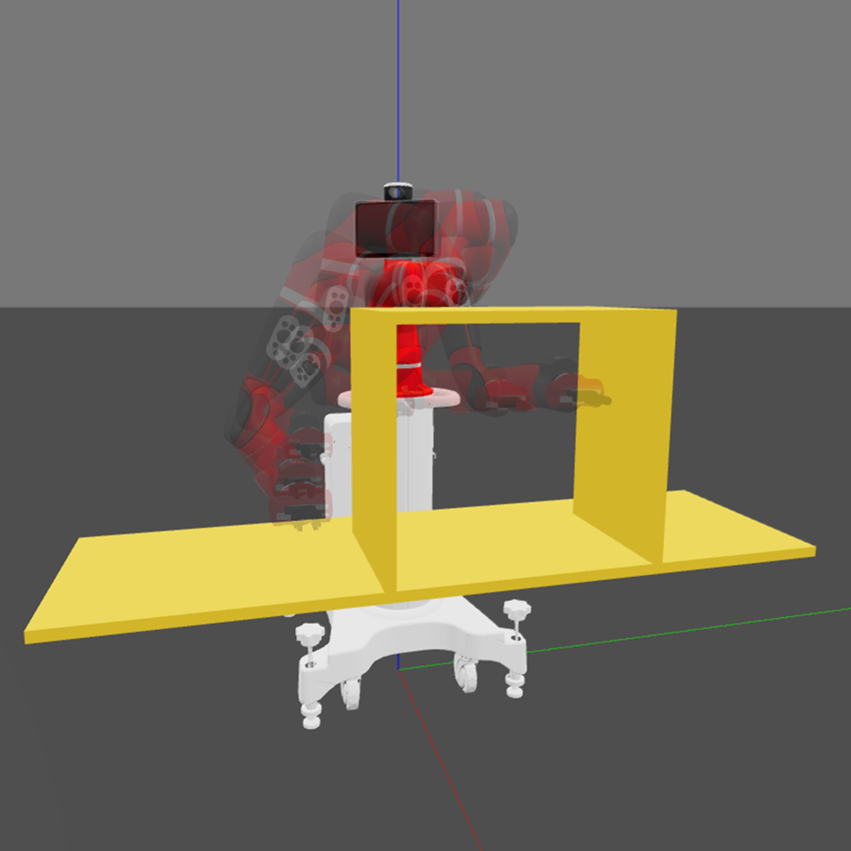}
    }
    \hfill
    \subfloat[Narrow Gap\label{fig:sub_d}]{
        \includegraphics[width=0.45\columnwidth]{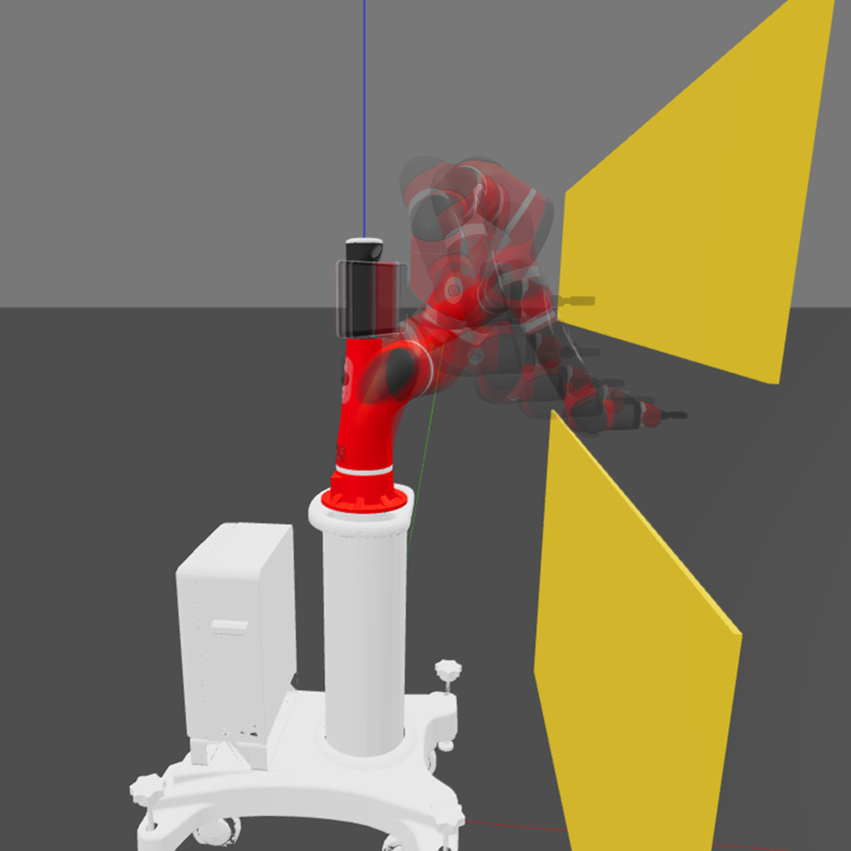}
    }
    
    \caption{Four different test scenes used for algorithm evaluation. Yellow objects denote obstacles. Overlaid robot configurations illustrate a successful motion sequence generated by our method in each scene.}
    \label{fig:scenes}
\end{figure}

We evaluate the proposed method in four simulated scenes: Shelf, Counter, Industrial, and Narrow Gap, as shown in Fig.~\ref{fig:scenes}. The first three are adapted from \cite{9343718} and correspond to representative challenging environments from the MoveIt! benchmark. The Narrow Gap scene is based on Scenario 2 in \cite{10802215} and is designed to highlight the effectiveness of our null-space obstacle avoidance strategy. For each scene, we construct 50 problem instances with different start and goal positions, denoted by \(\boldsymbol{p}_{\mathrm{start}}\) and \(\boldsymbol{p}_{\mathrm{goal}}\), respectively. 
Performance is evaluated using the following five metrics:
\begin{enumerate}[label=(\roman*)]
\item Trajectory tracking success rate \(SR\). A trial is counted as successful if the algorithm generates a collision-free task-space reference trajectory and the manipulator tracks it to completion without collision or joint limit violation. The success rate is defined as the ratio of successful executions to all trials for which a collision-free reference trajectory is found, and is used to measure overall algorithm performance.
\item Integrated position error.
\[
E_\mathrm{p}=\int_{0}^{T} \|\boldsymbol{p}_e(t)\|\, dt, 
\]
where $\boldsymbol{p}_e(t)$ is defined in Eq.~\eqref{eq:position_error}.
\item Integrated tool-axis orientation error. 
\[E_{\mathrm{z}}=\int_{0}^{T} \arccos(\boldsymbol{z}_{r}(t) \cdot \boldsymbol{z}(t))\,dt,
\]
where $\boldsymbol{z}_{r}(t)$ and $\boldsymbol{z}(t)$ are defined in Section~\ref{sec:Notation and Problem Formulation}. 
\item Integrated slosh-free error angle.
\[E_{\mathrm{sf}}=\int_{0}^{T} e_{\mathrm{sf}}(t) \,dt \in \mathbb{R}.\]
\item Maximum slosh-free error angle.
\[\max e_{\mathrm{sf}} = \max_{t \in [0, T]} e_{\mathrm{sf}}(t).\]
\end{enumerate}

The metrics \(E_\mathrm{p}\) and \(E_{\mathrm{z}}\) quantify tracking accuracy, with lower values indicating better tracking performance. The metrics \(E_{\mathrm{sf}}\) and \(\max e_{\mathrm{sf}}\) quantify how well a trajectory satisfies the nominal slosh-free condition, and are used as indirect measures of slosh-free
performance \cite{10610813}. Specifically, \(E_{\mathrm{sf}}\) reflects the accumulated deviation from the nominal slosh-free condition over the entire trajectory, while \(\max e_{\mathrm{sf}}\) captures the worst-case deviation. Lower values indicate closer adherence to slosh-free motion.

\subsubsection{Trajectory and controller parameters}

Task-space position trajectories are parameterized by degree-\(p\) B-splines with \(p=6\), using \(n+1=22\) control points and a knot vector of length \(m+1=29\). To ensure smooth liquid transport, we impose bounds on higher-order kinematic quantities: a velocity limit of \(v_{\text{max}} = 0.2 \,\mathrm{m/s}\), an acceleration limit of \(a_{\text{max}} = 1.0 \,\mathrm{m/s^2}\), a jerk limit of \(j_{\text{max}} = 3.0 \,\mathrm{m/s^3}\), and a snap limit of \(s_{\text{max}} = 3.0 \,\mathrm{m/s^4}\). For slosh-free orientation generation, the auxiliary constant is set to \(\sigma=\pi/2\).

The controller parameters are set as follows. In Eq.~(\ref{dq_cmd_wo_nullspace}), the damping factor in DLS pseudoinverse is \(\lambda=0.001\). In Eq.~\eqref{eq:xd}, the feedforward gain is \(k_v=1\). Since position and orientation errors play different roles in control, we use separate feedback gains: \(k_{e,p}=1.0\) for position error and \(k_{e,\psi}=2.0\) for orientation error. In Eq.~(\ref{safe_distance_env}), the environment-collision cutoff margin is set to \(\delta^{\text{env}}=0.3 \,\mathrm{m}\), and in Eq.~(\ref{safe_distance_self}), the self-collision safety margin is set to \(\delta^{\text{self}}=0.1 \,\mathrm{m}\). 
The main adjusted parameters are the null-space gain \(\alpha_{\text{null}}\) and the subtask weights $w_k$ in Eq.~(\ref{eq:combined_control:a}). The values of (\(\alpha_{\text{null}}, w_{\text{env}}, w_{\text{joint}}\)) are set to (\(0.15,4.0,3.0\)), (\(0.20,3.0,3.0\)), (\(0.10,3.0,3.0\)), and (\(0.15,4.0,1.0\)) for the Shelf, Counter, Industrial, and Narrow Gap scenes, respectively, while the self-collision weight is fixed as \(w_{\text{self}}=1.0\) in all scenes.

To facilitate the practical use of these parameters, we further conducted a parameter-sensitivity study with respect to \(\alpha_{\text{null}}\), \(w_{\text{env}}\), and \(w_{\text{joint}}\). For each scene, three candidate values were tested for each parameter, resulting in 27 controller configurations. The same reference trajectories and initial joint configurations were used for all configurations, and the planning stage was excluded from this comparison. The detailed experimental setup and statistical results are provided in Supplementary Material S3. The results show that \(\alpha_{\text{null}}\) and \(w_{\text{env}}\) are the dominant sensitive parameters, whereas \(w_{\text{joint}}\) has only a minor influence on the tracking performance. Increasing \(\alpha_{\text{null}}\) generally strengthens the secondary-task correction but may increase the position-tracking error, indicating a trade-off between secondary-task execution and primary-task tracking accuracy.

Based on these observations, we recommend starting from a moderate setting, e.g., \(w_{\text{env}}=3\), \(w_{\text{joint}}=2\), and \(\alpha_{\text{null}}=0.15\), for a new scene. During tuning, \(w_{\text{joint}}\) can usually be kept fixed at a moderate value unless the robot frequently approaches joint limits. The main tuning should focus on \(w_{\text{env}}\) and \(\alpha_{\text{null}}\). Specifically, \(w_{\text{env}}\) can be increased when the robot operates closer to obstacles or when collision-related failures are observed. If increasing \(w_{\text{env}}\) provides only limited improvement in obstacle avoidance, this may indicate that the overall null-space correction is insufficient. In this case, \(\alpha_{\text{null}}\) can be gradually increased. Nevertheless, \(\alpha_{\text{null}}\) and \(w_{\text{env}}\) should be adjusted cautiously because they are sensitive with respect to tracking accuracy. A practical systematic procedure is to perform a small-scale sweep over \(w_{\text{env}}\) and \(\alpha_{\text{null}}\) using a few representative trajectories, discard configurations with low success rates, and then select the parameter configuration with relatively small weights and gain \(\alpha_{\text{null}}\) that still satisfies the safety and tracking requirements.

\subsubsection{FeasibilityNet training details}

FeasibilityNet is implemented in PyTorch. The dataset is collected in the Counter scene and contains \(K=165\) reference trajectories. For each trajectory, up to  \(M_t=20\) initial joint configurations are sampled, resulting in a total of 3218 trajectory–configuration samples. The dataset is split into training and validation sets with a ratio of 8:2. For label generation, we set \(\eta_{\text{self}}=0.2\), \(\eta_{\text{joint}}=1.0\), \(\alpha=0.4\), and \(\beta=0.5\). During deployment, FeasibilityNet selects the highest-scoring configuration from \(M=21\) candidates. The first 20 candidates are obtained by solving inverse kinematics from random initial guesses, whereas the last one is obtained from a zero initial guess. FeasibilityNet evaluated all 21 candidate initial configurations in a single batch with an average inference time of 0.192 ms, measured over 100 forward passes on the NVIDIA GeForce RTX 4080 GPU. To evaluate cross-scene generalization, FeasibilityNet is trained only on the Counter scene and then directly deployed in the other three scenes. Training starts with an initial learning rate of 0.001. A linear warm-up schedule is used for the first 50 epochs, followed by cosine annealing. The network is trained with mean squared error loss, a batch size of 128, and a total of 1000 epochs.

\begin{table} 
\centering
\caption{Performance comparison of the proposed method and its variants across five evaluation metrics. Units: SR\,(\%), $E_\mathrm{p}\,(\mathrm{mm} \cdot \mathrm{s})$,   $E_{\mathrm{z}}\,(\mathrm{deg} \cdot \mathrm{s})$, $E_{\mathrm{sf}}\,(\mathrm{deg} \cdot \mathrm{s})$, $\max e_{\mathrm{sf}}\,(\mathrm{deg})$. $\downarrow$ indicates lower is better, $\uparrow$ indicates higher is better. Boldface indicates the best mean value for each metric. If two or more methods have the same mean value, the one with the smaller standard deviation is highlighted. }
\label{tab:comparison}
\setlength{\tabcolsep}{1.6pt} 
\resizebox{\linewidth}{!}{%
\begin{tabular}{clcccccccccc}
\toprule
 && \multicolumn{5}{c}{Shelf} & \multicolumn{5}{c}{Counter} \\
\cmidrule(lr){3-7} \cmidrule(lr){8-12} 
 ID & Method & SR\,(\%)$\uparrow$  & $E_\mathrm{p}$$\downarrow$ & $E_{\mathrm{z}}$$\downarrow$ & $E_{\mathrm{sf}}$$\downarrow$ & max $e_{\mathrm{sf}}$$\downarrow$   & SR\,(\%)$\uparrow$ & $E_\mathrm{p}$$\downarrow$ & $E_{\mathrm{z}}$$\downarrow$ & $E_{\mathrm{sf}}$$\downarrow$ & max $e_{\mathrm{sf}}$ $\downarrow$ \\
\midrule
M1 &Baseline &62.79 & 26.46 (27.61) & \textbf{0.129} (0.103) &3.153 (0.719) & 1.678 (0.721) & 64.29 & 63.05 (87.94) & 0.386 (0.624) & 4.125 (1.398) & 1.887 (1.320) \\
M2 &SF &63.64 & 28.36 (27.95) &0.155 (0.106) &0.397 (0.453) &0.518 (0.836) &63.16 &57.04 (75.21) &0.367 (0.423) & 0.733 (0.910) & \textbf{0.679} (1.633) \\
M3 &FN & 76.74 & 27.15 (28.48) & 0.136 (0.110) & 3.164 (0.693) &1.617 (0.827) &92.31 & \textbf{35.34} (52.37) & \textbf{0.225} (0.287) &4.002 (0.955) & 1.726 (0.951)\\
M4 &SF-FN & 74.42 & \textbf{25.46} (28.11) & 0.146 (0.101) & 0.329 (0.428) & 0.376 (0.731) &\textbf{100.00} & 45.79 (64.67) & 0.280 (0.347) &0.648 (0.807) & 0.865 (1.809)  \\
M5 &NS & 63.64 & 40.71 (9.80) & 0.138 (0.068) & 3.161 (0.685) &1.387 (0.163) & 71.79 & 71.19 (69.87) &0.343 (0.384) & 4.125 (1.157) & 1.939 (1.061) \\
M6 &NS-SF & 67.44 & 40.76 (9.78) &0.147 (0.068) & \textbf{0.241} (0.137) & \textbf{0.138} (0.150) & 65.00 & 74.11 (75.45) &0.372 (0.423) &0.765 (0.874) & 0.693 (1.669) \\
M7 &NS-FN & \textbf{82.93} & 40.19 (11.57) & 0.136 (0.073) & 3.161 (0.718) &1.395 (0.242) &92.31 & 50.59 (51.14) & 0.234 (0.332) &4.050 (0.964) &1.909 (1.333)\\
M8 &NS-SF-FN (IPSO)&81.82 & 41.89 (11.98) &0.160 (0.085) &0.265 (0.180) & 0.184 (0.267) & \textbf{100.00} & 62.20 (63.26) & 0.303 (0.368) &\textbf{0.621} (0.783) & 0.686 (1.668) \\
\midrule[\heavyrulewidth] 

    && \multicolumn{5}{c}{Industrial} & \multicolumn{5}{c}{Narrow Gap} \\
\cmidrule(lr){3-7} \cmidrule(lr){8-12} 
ID & Method & SR\,(\%)$\uparrow$ & $E_\mathrm{p}$$\downarrow$ & $E_{\mathrm{z}}$$\downarrow$ & $E_{\mathrm{sf}}$$\downarrow$ & max $e_{\mathrm{sf}}$$\downarrow$   & SR\,(\%)$\uparrow$ & $E_\mathrm{p}$$\downarrow$ & $E_{\mathrm{z}}$$\downarrow$ & $E_{\mathrm{sf}}$$\downarrow$ & max $e_{\mathrm{sf}}$$\downarrow$  \\
\midrule
M1 &Baseline        &73.33 &10.88 (5.46) &0.069 (0.038) &2.565 (0.945) &1.215 (0.294) &57.14 &8.92 (3.72) &\textbf{0.063} (0.032) &1.737 (0.443) &1.169 (0.198) \\
M2 &SF              &72.34 &11.40 (6.01) &0.086 (0.039) &0.140 (0.085) &0.090 (0.114) &55.10 &8.95 (3.79) &0.080 (0.033) &0.102 (0.038) &0.043 (0.022) \\
M3 &FN              &95.65 &10.81 (4.98) &\textbf{0.062} (0.034) &2.784 (1.028) &1.287 (0.300) &64.00 &\textbf{8.85} (3.68) &0.065 (0.032) &1.817 (0.398) &1.201 (0.179) \\
M4 &SF-FN           &\textbf{97.78} &\textbf{10.54} (4.68) &0.076 (0.026) &\textbf{0.117} (0.056) &\textbf{0.061} (0.065) &62.00 &8.87 (3.75) &0.082 (0.033) &\textbf{0.101} (0.037) &0.039 (0.016) \\
M5 &NS              &72.34 &21.82 (6.47) &0.097 (0.046) &2.585 (0.938) &1.218 (0.260) &71.43 &32.02 (4.87) &0.109 (0.048) &1.758 (0.429) &1.184 (0.203) \\
M6 &NS-SF           &73.91 &21.88 (6.64) &0.108 (0.049) &0.161 (0.098) &0.091 (0.125) &72.00 &32.20 (5.12) &0.118 (0.048) &0.134 (0.043) &0.043 (0.017) \\
M7 &NS-FN           &97.67 &20.61 (4.87) &0.076 (0.040) &2.659 (0.919) &1.245 (0.279) &\textbf{75.51} &30.70 (5.71) &0.101 (0.052) &1.803 (0.440) &1.189 (0.194) \\
M8 &NS-SF-FN(IPSO)&95.65 &21.43 (5.46) &0.096 (0.042) &0.142 (0.070) &0.070 (0.078) &74.00 &28.25 (5.66) &0.112 (0.049) &0.127 (0.048) &\textbf{0.039} (0.012) \\
\bottomrule
\end{tabular}%
}
\end{table}

\subsection{Ablation study}
This section presents an ablation study of three key components of IPSO: slosh-free orientation generation in Module 1, initial joint configuration selection in Module 2, and null-space correction in Module 3. By enabling or disabling these components in different combinations, we define eight methods for comparison, as summarized in Table~\ref{tab:comparison}.

In the method names, SF indicates that the reference trajectory is constructed using the slosh-free orientation generation strategy introduced in Section~\ref{sec:Generation of Slosh-free Reference Trajectories in Cluttered Environments}. If SF is absent, the reference orientation is kept constant throughout the motion and is set to \(\boldsymbol{R}_{r0}\), the orientation given by Eq.~(\ref{eq:sf orientation}) at \(t=0\). FN indicates that the initial joint configuration is selected using the learning-based strategy proposed in Section~\ref{sec:Learning-Based Evaluation and Selection of Initial Joint Configurations}; otherwise, the initial joint configuration is obtained by solving inverse kinematics from the zero joint configuration. NS indicates that the null-space correction term in the tracking controller (Eq.~(\ref{eq:rrmc_nullspace})) is enabled, i.e., \(\alpha_{\text{null}} > 0\); otherwise, \(\alpha_{\text{null}} = 0\). Here, Baseline denotes the reference method in which all three components are replaced, whereas NS-SF-FN denotes the complete method proposed in Section~\ref{method}, i.e., IPSO.

The eight methods are evaluated in four simulated scenes, and the results are reported in Table~\ref{tab:comparison}. Except for the success rate, all metrics are presented as mean (standard deviation).

\subsubsection{Effect of slosh-free orientation generation}
To evaluate the effect of slosh-free orientation generation, we compare all method pairs that differ only in whether SF is enabled, namely M1--M2, M3--M4, M5--M6, and M7--M8. Across all scenes, enabling SF consistently reduces both \(E_{\mathrm{sf}}\) and \(\max e_{\mathrm{sf}}\). The corresponding values without SF are approximately 5–24 and 2–31 times those obtained with SF, respectively. These results show that generating the reference orientation according to Eq.~(\ref{eq:sf orientation}) substantially improves adherence to the slosh-free condition and provides indirect evidence of improved sloshing suppression. This effect remains consistent across scenes and combinations of the other components.

\subsubsection{Effect of initial joint configuration selection}

To evaluate the effect of the initial joint configuration selection component, we compare the method pairs that differ only in whether FN is enabled: M1--M3, M2--M4, M5--M7, and M6--M8. In the Shelf, Counter, and Industrial scenes, enabling FN improves the success rate by 10.78--19.29, 20.52--36.84, and 21.74--25.44 percentage points, respectively.

By contrast, the improvement in the Narrow Gap scene is smaller, ranging from 2.00 to 6.90 percentage points. In the tested tasks, the manipulator performs a relatively short motion through the passage, mainly by extending the arm outward from one side of the gap to the other. Such motions are less likely to drive the robot toward self-collision or joint-limit violations, which are the two main risks considered in the supervision of FeasibilityNet. Consequently, the selected initial configuration has less influence on execution success in this scene. Nevertheless, enabling FN does not reduce the success rate in any of the four pairwise comparisons, indicating that it avoids selecting clearly unfavorable initial configurations even when the room for its improvement is limited.

Beyond the Counter scene used for training-data collection, FeasibilityNet also improves the success rate in the other three unseen scenes, although the improvement is relatively limited in Narrow Gap. These results provide evidence of cross-scene generalization within the evaluated environments, which is further supported by the generalization analysis in Supplementary Material S4. Overall, the proposed selection strategy helps filter out unfavorable initial configurations before execution, thereby improving the reliability of long-horizon trajectory tracking.

However, this cross-scene performance does not imply that FeasibilityNet learns generic kinematic features of the Sawyer robot. Because its supervision is generated from trajectory-tracking rollouts under a fixed robot model and downstream controller, the network learns robot- and controller-conditioned compatibility patterns between an initial joint configuration and a reference trajectory. The environment topology is not explicitly provided to the network, but influences its prediction indirectly through the distribution of reference trajectories generated by Module 1. Consequently, environments with substantially different topologies may produce out-of-distribution trajectory--configuration pairs and degrade prediction performance.

We further quantified the computational cost of generating the offline supervision used to train FeasibilityNet. For 200 sampled start--goal pairs, the data-generation process produced 165 valid reference trajectories and 3,218 trajectory--configuration samples, requiring approximately 3.60~h in total. Including unsuccessful planning attempts, the amortized reference-trajectory generation time was 5.93~s per valid trajectory, while generating a batch of 20 candidate joint configurations via inverse kinematics (IK) required $0.290\pm0.159$~s per trajectory. By comparison, rollout-based labeling required $3.710\pm1.504$~s per trajectory--configuration pair and accounted for approximately 92.0\% of the total data-generation time. Therefore, the main cost of extending the dataset to additional scenes arises from rollout-based supervision rather than trajectory or IK-candidate generation.

Given this additional labeling cost, we further examined the benefit of multi-scene training by training another FeasibilityNet on the combined Counter and Shelf datasets while keeping the architecture and training settings unchanged. As detailed in Supplementary Material S4, multi-scene training exposed the network to a broader distribution of trajectory--configuration pairs and improved score approximation and ranking performance in most evaluated environments. However, these improvements did not consistently translate into higher overall task success for IPSO. This result suggests that additional multi-scene data can improve the prediction model itself, but effectively converting this improvement into task-level gains may require task-aligned ranking objectives, scene-balanced sampling, and out-of-distribution detection.

\subsubsection{Effect of null-space correction}

To evaluate the effect of null-space correction, we compare all method pairs that differ only in whether NS is enabled, namely M1--M5, M2--M6, M3--M7, and M4--M8. The largest and most consistent success rate improvement is observed in the Narrow Gap scene, whereas the effect is limited in the other three scenes.

This difference is closely related to the obstacle layout in each scene. In Shelf, Counter, and Industrial, the obstacles primarily constrain the end-effector path while leaving relatively more clearance for the manipulator links; therefore, the collision-free task-space reference trajectory already addresses most obstacle-avoidance requirements. In Narrow Gap, the obstacles also restrict the manipulator links, making task-space planning alone insufficient for collision-free execution. Under this condition, NS improves success by adjusting the manipulator posture through its redundant degrees of freedom.

Enabling NS generally increases \(E_\mathrm{p}\) and, in most comparisons,  \(E_\mathrm{z}\), indicating some degradation in primary-task tracking accuracy. However,  \(E_{\mathrm{sf}}\) and \(\max e_{\mathrm{sf}}\) remain largely unchanged, showing that the additional safety correction does not noticeably compromise adherence to the slosh-free condition.

Taken together, the ablation results show that the three components play distinct but complementary roles in the overall framework. SF primarily serves the objective of slosh-free motion. Across all combinations of FN and NS, enabling SF consistently reduces \(E_{\mathrm{sf}}\) and \(\max e_{\mathrm{sf}}\) (M1–M2, M3–M4, M5–M6, and M7–M8), while maintaining broadly comparable success rates. In contrast, FN and NS mainly improve the safe joint-space realization of the reference motion, but act at different stages. FN selects a trajectory-compatible initial joint-space branch before execution, whereas NS provides local safety correction online. Their relative contributions are scene dependent (comparing M1, M3, M5, and M7 without SF, and M2, M4, M6, and M8 with SF). For example, FN provides the dominant success rate improvement in Shelf and Industrial, whereas NS is more important in Narrow Gap. Nevertheless, combining FN and NS (M7 and M8) generally achieves a success rate comparable to or higher than using either component alone (M3/M5 and M4/M6, respectively), indicating that favorable initialization and online correction address different failure modes rather than being redundant. Therefore, the complete framework combines slosh-free reference generation, trajectory-conditioned initialization, and local safety correction to address different requirements of the same transport task.

\subsection{Comparison with external baselines}
To further evaluate the proposed method in liquid-transport tasks involving multiple constraints, we compare it with two external baselines representing a unified optimization approach and a cascaded trajectory-tracking approach, respectively. Specifically, the baselines include an MPC method adapted for liquid transportation and the geometric slosh-free tracking (GSFT) method.

\subsubsection{MPC for liquid transportation}
We construct the MPC baseline based on the method proposed in \cite{10285028} and its publicly available lightweight implementation.\footnote{Available at:
\url{https://github.com/adamheins/uplite}} Although it was not developed specifically for liquid transportation, it is closely related to the spill-free transport problem considered here. We extend the original implementation to account for environmental collisions and robot self-collisions. The collision-avoidance and spill-free requirements are incorporated into the objective function as penalties to prevent the MPC optimization problem from becoming infeasible when multiple constraints must be satisfied simultaneously. This baseline represents a holistic optimization-based approach that handles motion generation, control, and multiple task constraints within a single optimization framework. Its comparison with our integrated planning-and-control framework evaluates whether decomposing and separately handling the constraints improves task performance.

The resulting baseline performs online liquid transportation by solving the following MPC problem:
\begin{equation}
\begin{aligned}
\min_{\substack{
\boldsymbol{x}_{t:t+N},\\
\boldsymbol{u}_{t:t+N-1}}}
\quad &
\sum_{k=t}^{t+N-1}
\ell\!\left(\boldsymbol{x}_k,\boldsymbol{u}_k\right)
+
\ell_N\!\left(\boldsymbol{x}_{t+N}\right)
\\
\mathrm{s.t.}\quad
&
\boldsymbol{x}_{k+1}
=
\boldsymbol{F}_{\Delta t}
\!\left(\boldsymbol{x}_k,\boldsymbol{u}_k\right),
\quad k=t,\ldots,t+N-1,
\\
&
\boldsymbol{x}_t=\boldsymbol{x}(t),
\\
&
\underline{\boldsymbol{x}}
\leq
\boldsymbol{x}_k
\leq
\overline{\boldsymbol{x}},
\quad k=t,\ldots,t+N,
\\
&
\underline{\boldsymbol{u}}
\leq
\boldsymbol{u}_k
\leq
\overline{\boldsymbol{u}},
\quad k=t,\ldots,t+N-1.
\end{aligned}
\label{eq:mpc_baseline}
\end{equation}
Here,
$\boldsymbol{x}_k=
[\boldsymbol{q}_k^\top,
 \dot{\boldsymbol{q}}_k^\top,
 \ddot{\boldsymbol{q}}_k^\top]^\top$
denotes the joint-space state, and
$\boldsymbol{u}_k=\dddot{\boldsymbol{q}}_k$
denotes the joint jerk input.
The parameter $N$ is the prediction horizon in each MPC optimization, while $t$
denotes the current discrete time step. The mapping
$\boldsymbol{F}_{\Delta t}$ denotes the discrete state-transition
model with sampling interval $\Delta t$. Specifically, a joint-space
triple-integrator model is adopted:
\[
\frac{\mathrm{d}}{\mathrm{d}t}
\begin{bmatrix}
\boldsymbol{q}\\
\dot{\boldsymbol{q}}\\
\ddot{\boldsymbol{q}}
\end{bmatrix}
=
\begin{bmatrix}
\dot{\boldsymbol{q}}\\
\ddot{\boldsymbol{q}}\\
\boldsymbol{u}
\end{bmatrix}.
\]
The vectors
$\underline{\boldsymbol{x}}$ and $\overline{\boldsymbol{x}}$ denote
the lower and upper state bounds, respectively, whereas
$\underline{\boldsymbol{u}}$ and $\overline{\boldsymbol{u}}$ denote
the lower and upper input bounds. The functions
$\ell(\boldsymbol{x}_k,\boldsymbol{u}_k)$ and
$\ell_N(\boldsymbol{x}_{t+N})$ denote the stage cost and terminal cost,
respectively.
\begin{subequations}
\label{eq:mpc_costs}
\begin{align}
\ell(\boldsymbol{x}_k,\boldsymbol{u}_k)
&=
\left\|
\Delta\boldsymbol{p}_k
\right\|_{\boldsymbol{W}_{p}}^{2}
+
\left\|
\boldsymbol{x}_k
\right\|_{\boldsymbol{W}_{x}}^{2}
+
\left\|
\boldsymbol{u}_k
\right\|_{\boldsymbol{W}_{u}}^{2}
\notag
\\
&\quad+
\left\|
\boldsymbol{\rho}_{\mathrm{obs}}(\boldsymbol{x}_k)
\right\|_{\boldsymbol{W}_{\mathrm{obs}}}^{2}
+
\left\|
\boldsymbol{\rho}_{\mathrm{sf}}(\boldsymbol{x}_k)
\right\|_{\boldsymbol{W}_{\mathrm{sf}}}^{2}
\notag
\\
&\quad+
\left\|
\boldsymbol{\rho}_{\mathrm{self}}(\boldsymbol{x}_k)
\right\|_{\boldsymbol{W}_{\mathrm{self}}}^{2},
\label{eq:mpc_stage_cost}
\\
\ell_N\!\left(\boldsymbol{x}_{t+N}\right)
&=
\left\|
\Delta\boldsymbol{p}_{t+N}
\right\|_{\boldsymbol{W}_{p,N}}^{2}
+
\left\|
\boldsymbol{x}_{t+N}
\right\|_{\boldsymbol{W}_{x,N}}^{2}.
\label{eq:mpc_terminal_cost}
\end{align}
\end{subequations}

Here, $\|\boldsymbol{\xi}\|_{\boldsymbol{W}}^{2}
= \boldsymbol{\xi}^{T}\boldsymbol{W}\boldsymbol{\xi}$,
where the $\boldsymbol{W}$ terms are positive semidefinite weighting matrices. The end-effector position error is defined as $\Delta\boldsymbol{p}_k = \boldsymbol{p}_{\mathrm{ee}}(\boldsymbol{q}_k) -
\boldsymbol{p}_{d}$, where $\boldsymbol{p}_{d}$ is the desired end-effector position provided as an input to the MPC problem.
The terms $\boldsymbol{\rho}_{\mathrm{obs}}$,
$\boldsymbol{\rho}_{\mathrm{sf}}$, and
$\boldsymbol{\rho}_{\mathrm{self}}$ denote the environmental collision, slosh-free alignment, and self-collision penalty residuals, respectively. The obstacle and self-collision penalties are constructed using the same signed distance functions and collision-sphere representation as those described in Section~\ref{sec:Obstacle-Aware Tracking via Null-Space Projection}. A nonzero environmental-collision penalty is introduced when the obstacle clearance falls below the prescribed safety margin, while the self-collision penalty becomes active when a selected pair of collision spheres overlaps. The spill-free residual penalizes misalignment between the $z$-axis of the end-effector tool frame and the apparent-acceleration direction at the end effector. Further implementation details and the MPC parameter settings are provided in the Supplementary Material S5.

\subsubsection{GSFT}
GSFT \cite{10610813} is a geometric slosh-free tracking method specifically developed for liquid transportation tasks. Similar to our approach, GSFT adopts a cascaded trajectory generation and tracking framework. This comparison evaluates whether the proposed IPSO framework can extend an existing cascaded slosh-free tracking approach with collision avoidance while preserving spill-free transportation performance.

To ensure a fair comparison, the same slosh-free end-effector reference trajectories are used for both methods. Specifically, all test trajectories are generated by the trajectory generation module of our method in Section~\ref{sec:Generation of Slosh-free Reference Trajectories in Cluttered Environments} and satisfy both collision-free and slosh-free requirements. For the GSFT baseline, we adopt the control framework provided in its publicly available implementation\footnote{Available at:
\url{https://github.com/jonarriza96/gsft}}, including a cascaded proportional-derivative task-space controller and a joint-space controller based on convex quadratic programming. The original control parameters are kept unchanged.
\begin{table}
\centering
\caption{Performance comparison of different methods across four scenes. Units: SR\,(\%), $E_{\mathrm{sf}}\,(\mathrm{deg}\cdot\mathrm{s})$, max $e_{\mathrm{sf}}\,(\mathrm{deg})$. The values in parentheses indicate standard deviations.}
\label{tab:comparison_methods}
\setlength{\tabcolsep}{1.0pt}

\begin{tabular}{lccc ccc}
\toprule
& \multicolumn{3}{c}{Shelf}
& \multicolumn{3}{c}{Counter}
\\

\cmidrule(lr){2-4}
\cmidrule(lr){5-7}

Method
& SR\,(\%)$\uparrow$
& $E_{\mathrm{sf}}$$\downarrow$
& max $e_{\mathrm{sf}}$$\downarrow$

& SR\,(\%)$\uparrow$
& $E_{\mathrm{sf}}$$\downarrow$
& max $e_{\mathrm{sf}}$$\downarrow$
\\

\midrule

IPSO &81.82 &0.169(0.066) &0.051(0.011) &100.00 &-- &-- \\
MPC  &18.18 &11.482(4.460) &3.904(1.571) &0.00 &-- &-- \\

\midrule

IPSO &81.82 &0.264(0.190) &0.193(0.296) &100.00 &0.346(0.459) &0.569(1.950) \\
GSFT &63.64 &0.413(0.722) &1.182(2.071) &47.37 &0.381(1.007) &1.189(2.029) \\

\midrule[\heavyrulewidth]

& \multicolumn{3}{c}{Industrial}
& \multicolumn{3}{c}{Narrow Gap}
\\

\cmidrule(lr){2-4}
\cmidrule(lr){5-7}

Method
& SR\,(\%)$\uparrow$
& $E_{\mathrm{sf}}$$\downarrow$
& max $e_{\mathrm{sf}}$$\downarrow$

& SR\,(\%)$\uparrow$
& $E_{\mathrm{sf}}$$\downarrow$
& max $e_{\mathrm{sf}}$$\downarrow$
\\

\midrule

IPSO &95.65 &0.087(0.044) &0.067(0.075) &74.00 &0.110(0.044) &0.040(0.016) \\
MPC  &10.87 &8.342(2.886) &3.275(1.354) &44.00 &5.718(2.713) &2.278(1.118) \\

\midrule

IPSO &95.65 &0.132(0.059) &0.055(0.051) &74.00 &0.121(0.052) &0.035(0.007) \\
GSFT &67.39 &0.099(0.233) &0.239(0.758) &56.00 &0.041(0.026) &0.095(0.316) \\

\bottomrule
\end{tabular}
\end{table}
\subsubsection{Results}The comparison results are reported in Table~\ref{tab:comparison_methods}. 
We compare IPSO with MPC and GSFT in terms of success rate (SR) and the slosh-free metrics $E_{\mathrm{sf}}$ and $\max e_{\mathrm{sf}}$. 
For the latter two metrics, the mean and standard deviation are computed only over the rollouts successfully completed by both methods in each pairwise comparison.

IPSO achieves the highest success rate in all four scenes, whereas MPC shows the lowest overall success rate and fails all trials in the Counter scene. Even among the rollouts successfully completed by both methods, the $E_{\mathrm{sf}}$ and $\max e_{\mathrm{sf}}$ values of MPC are generally more than one order of magnitude higher than those of IPSO. 
This indicates that, when joint limits, environmental collision avoidance, self-collision avoidance, and spill-free requirements are incorporated into a single online optimization problem, the coupling among multiple high-dimensional nonlinear constraints increases the difficulty of solving the problem online and makes the trade-offs among different objectives more challenging.

Compared with GSFT, IPSO achieves a higher task success rate in every scene while maintaining comparable liquid slosh-free performance. 
Although GSFT obtains a lower integrated error $E_{\mathrm{sf}}$ in some scenes, IPSO achieves lower $\max e_{\mathrm{sf}}$ values in all scenes and smaller standard deviations for most metrics. 
Because the GSFT controller does not explicitly account for environmental geometry, collisions may still occur during trajectory tracking even when the end-effector reference trajectory is collision-free, thereby reducing the task success rate. In contrast, IPSO improves motion reliability in complex environments while preserving slosh-free tracking performance comparable to GSFT.

Overall, the unified MPC formulation has difficulty handling multiple strongly coupled constraints, whereas GSFT provides effective slosh-free tracking but lacks explicit environmental constraint handling. 
By decomposing the problem into slosh-free trajectory generation, initial-configuration selection, and obstacle-aware trajectory tracking, IPSO reduces the coupling among different constraints and improves task completion in complex environments without sacrificing liquid stability.

\begin{figure}[pos=t]
\centering
\includegraphics[scale=0.4]{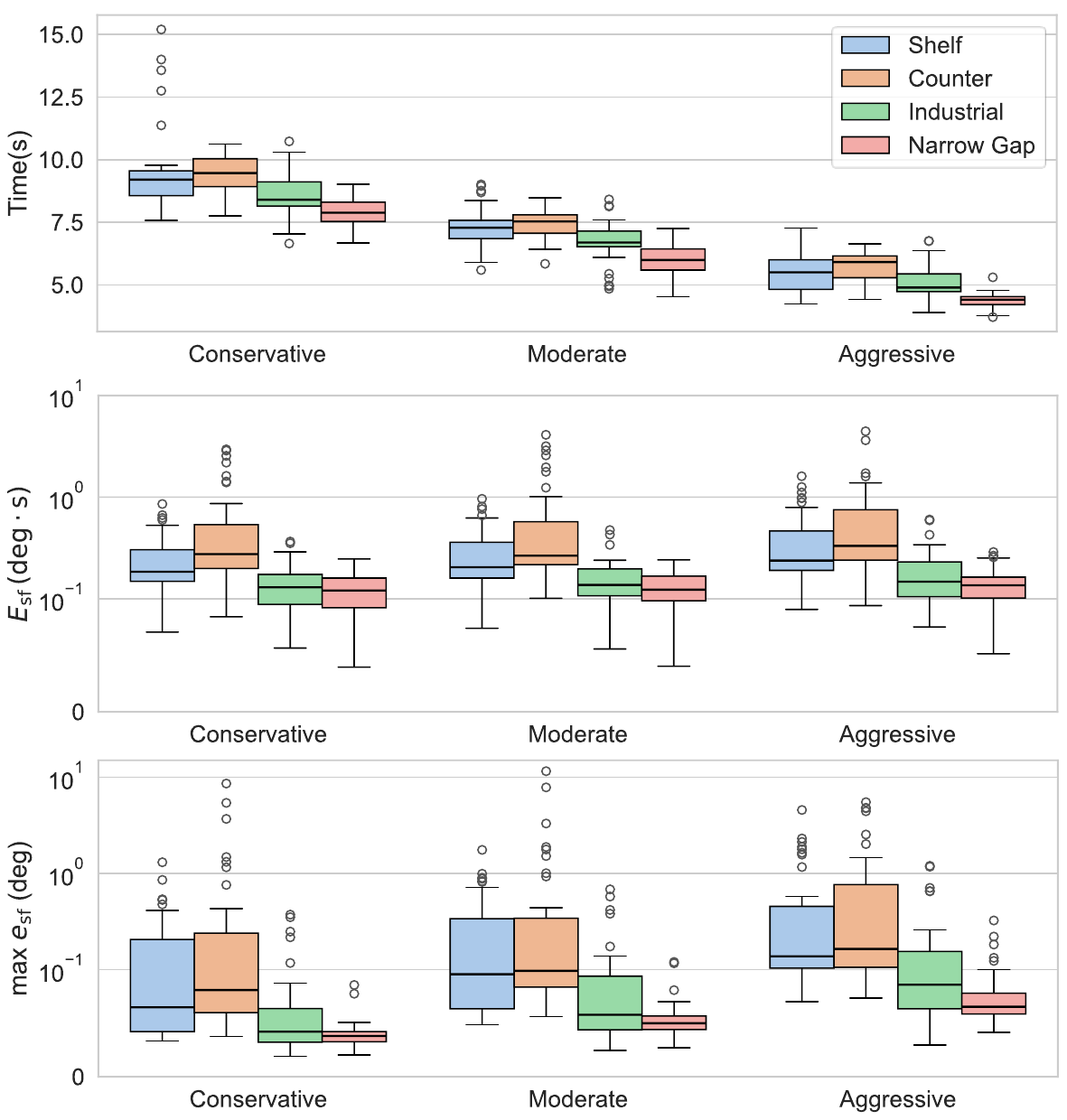} 
\caption{Comparison of trajectory duration \(T\), \(E_{\mathrm{sf}}\), and \(\max e_{\mathrm{sf}}\) under different kinematic constraint settings.}
\label{joint limit result}
\end{figure}

\subsection{Effect of kinematic constraints on trajectory
duration and nominal slosh-free tracking}
As shown in Table~\ref{tab:boundary_configs}, we consider three progressively relaxed kinematic-limit settings: Conservative, Moderate, and Aggressive. These settings impose different bounds on the velocity, acceleration, jerk, and snap of the reference trajectory, thereby producing different trajectory durations and tracking demands. For each scene and each setting, IPSO is evaluated on 50 planning problems. The statistics of the trajectory duration $T$, $E_{\mathrm{sf}}$, and $\max e_{\mathrm{sf}}$ in Fig.~\ref{joint limit result} are computed over successful trials only. The success rates remain broadly comparable across the three settings, so the observed trends are not primarily caused by changes in the number of successful trials.

Fig.~\ref{joint limit result} shows that, as the kinematic limits are relaxed from Conservative to Aggressive, the median trajectory duration decreases consistently in all four scenes. At the same time, the medians of $E_{\mathrm{sf}}$ and $\max e_{\mathrm{sf}}$ generally increase, with the most pronounced changes observed in Shelf and Counter. Since these two metrics quantify the deviation of the executed motion from the nominal slosh-free condition, the results show that faster reference motions can reduce execution time while making the nominal slosh-free constraint more difficult to track accurately. This result can be interpreted in light of the interaction
analysis in Section~\ref{sec: Module Interactions}, which shows that more aggressive time parameterization increases the translational tracking demand and can induce faster variation of the reference tool-axis direction. The observed increases in $E_{\mathrm{sf}}$ and $\max e_{\mathrm{sf}}$ indicate that, under the evaluated settings, these greater tracking demands
were accompanied by larger execution deviations from the nominal slosh-free condition.

Overall, the results indicate a scene-dependent trade-off between execution time and adherence to the nominal slosh-free condition. However, these results should not be interpreted as evidence that the assumptions underlying the nominal slosh-free condition have broken down, because $E_{\mathrm{sf}}$ and $\max e_{\mathrm{sf}}$ do not quantify the model-deviation angle $e_{\mathrm{qs}}$ or the actual liquid-surface response. In practice, more aggressive limits may be selected when execution time is prioritized, whereas more conservative limits are preferable when closer tracking of the nominal slosh-free condition is required.

\begin{table}[pos=t]
\centering
\caption{Kinematic limits under different settings.}
\label{tab:boundary_configs}
\begin{tabular}{lccc}
\toprule
Parameter & Conservative & Moderate & Aggressive \\
\midrule
$v_{\max}$ ($\mathrm{m/s}$)       & 0.2 & 0.5 & 1.0 \\
$a_{\max}$ ($\mathrm{m/s^2}$)     & 1.0 & 3.0 & 6.0 \\
$j_{\max}$ ($\mathrm{m/s^3}$)     & 3.0 & 6.0 & 12.0 \\
$s_{\max}$ ($\mathrm{m/s^4}$)     & 3.0 & 6.0 & 12.0 \\
\bottomrule
\end{tabular}
\end{table}

\begin{figure}[pos=t]
\centering
\includegraphics[width=\linewidth]{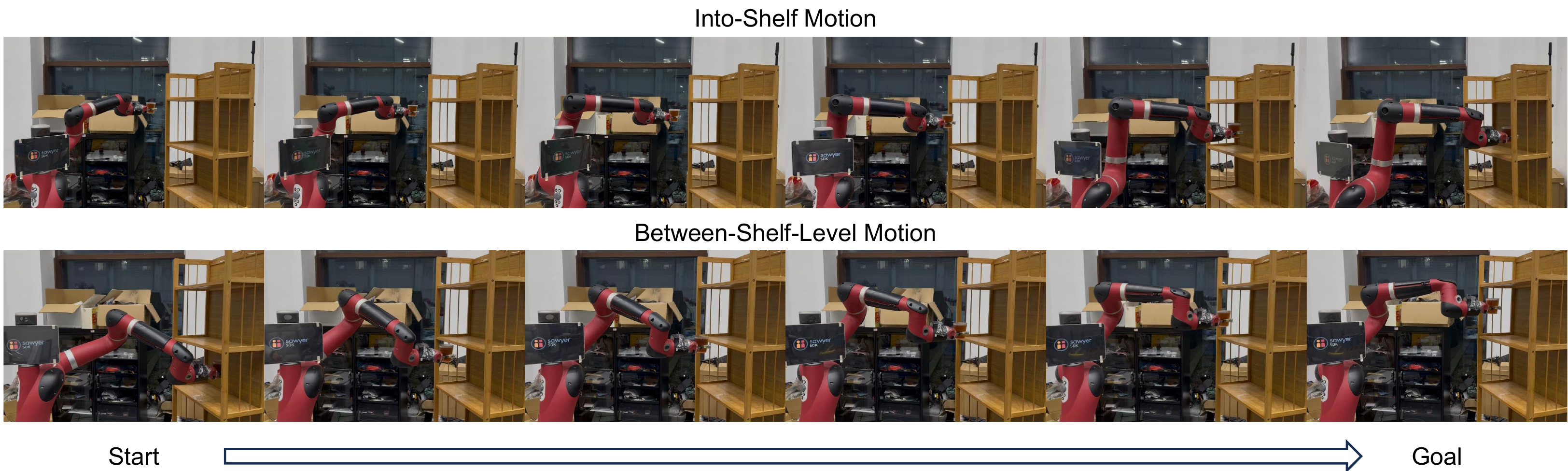} 
\caption{
Execution snapshots of spill-free liquid transport in a cluttered environment. The end-effector carries a round-rim cup filled to 90\% of its capacity with oolong tea. (Top) Into-Shelf motion. (Bottom) Between-Shelf-Level motion.}
\label{real-world exp}
\end{figure}

\begin{figure}[pos=t]
\centering
\includegraphics[scale=0.4]{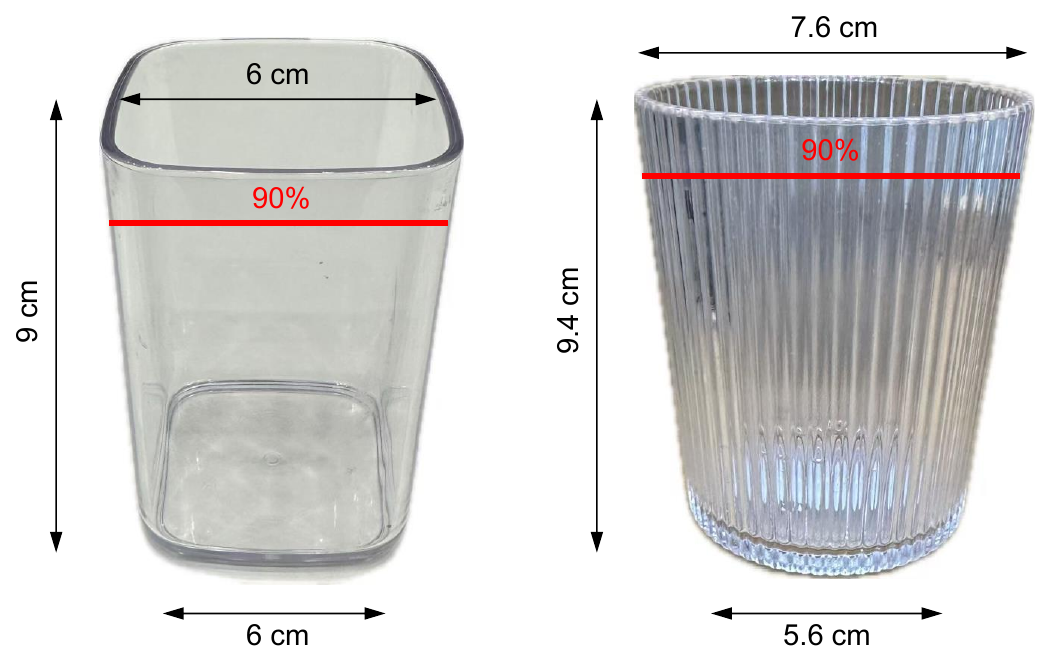} 
\caption{Shapes and dimensions of the two containers used in the real-world experiments: (left) a square-rim cup and (right) a round-rim cup. The red line indicates the approximate 90\% liquid level used for the corresponding high-fill transport experiments. }
\label{cup size}
\end{figure}

\subsection{Real-world experiments}

We validate the proposed method on a real 7-DoF Sawyer manipulator, using a bookshelf as the obstacle in the environment, as shown in Fig.~\ref{real-world exp}. The environment remains static throughout all real-world experiments, with no moving obstacles in the robot's workspace. Since this work focuses on liquid transport rather than how the container is grasped, the liquid-filled container is assumed to be securely grasped by the gripper before the task begins. Given a start and a goal end-effector position, our method first generates a reference trajectory and then executes the transport task by tracking that trajectory. Based on this setup, we conduct three types of real-world experiments:

\begin{enumerate}
    \item Collision-free liquid transport in a bookshelf environment. This experiment evaluates the overall capability of the proposed method in cluttered environments, namely whether it can achieve collision-free and spill-free liquid transport in the presence of obstacles. For each method, 10 executions are conducted. In 5 of them, the robot transports the container from outside the shelf to inside the shelf (Into-Shelf Motion). In the other 5, it moves the container between different shelf levels (Between-Shelf-Level Motion).
    \item Camera-based measurement of liquid oscillation. This experiment directly evaluates liquid-surface motion during transport using camera-based measurement. For each method, 10 executions are recorded and analyzed using the same reference trajectories as in the collision-free transport experiment, including 5 Into-Shelf Motion trials and 5 Between-Shelf-Level Motion trials.
    \item Transport with various containers and liquids. To examine applicability under different task conditions, we use two containers with different shapes (see Fig.~\ref{cup size}) and three everyday beverages: oolong tea, milk, and Coca-Cola, resulting in 6 container-liquid combinations. For each configuration, 5 Between-Shelf-Level Motion trials are performed.
\end{enumerate}

For the collision-free transport experiment and the experiments with diverse container--liquid combinations, the liquid volume is set to 90\% of the container capacity to provide a demanding transport condition. For the camera-based experiment, the cup is instead filled to approximately 60\% of its capacity to provide sufficient headspace for observable liquid-surface motion and visual measurement. The aggressive kinematic constraint setting is used in all three experiments.
\subsubsection{Collision-free liquid transport in a bookshelf environment}

We use a round-rim cup filled to 90\% of its capacity with oolong tea. During the experiments, the robot joint trajectories are recorded, and the corresponding \(E_{\mathrm{sf}}\) values are computed for all trials. These values quantify how well the executed motion satisfies the slosh-free condition and indirectly reflect slosh-free performance. The results are reported in Table~\ref{tab:Esf_rw}, where T0--T4 denote the trajectory indices for each motion type.

Using the complete method (M8), the robot successfully completed all 10 transport trials without any liquid spillage. Table~\ref{tab:Esf_rw} further compares M8 and M7, corresponding to the methods with the same IDs in Table~\ref{tab:comparison}. The two methods differ only in how the reference orientation trajectory is constructed: M8 uses a slosh-free reference orientation, whereas M7 uses a constant reference orientation. All other experimental settings are kept the same, so this comparison directly reflects the role of the slosh-free orientation generation module in the real system.

As shown in the table, M8 yields consistently lower \(E_{\mathrm{sf}}\) values than M7 for both motion types. Specifically, compared with M7, M8 reduces \(E_{\mathrm{sf}}\) by approximately 42\%--43\% for both Between-Shelf-Level and Into-Shelf motions. This indicates that the proposed method satisfies the slosh-free condition more closely and achieves better slosh-free performance. In addition, the smaller standard deviations obtained by M8 indicate less variation in  \(E_{\mathrm{sf}}\) across the evaluated trajectories. Representative execution snapshots of the two motion types are shown in Fig.~\ref{real-world exp}.

\subsubsection{Camera-based measurement of liquid oscillation}
To directly quantify liquid oscillation during real-robot transport, M8 and M7 were evaluated using a camera-based measurement procedure. Both methods executed the same ten reference trajectories using a square-rim cup. Liquid oscillation was quantified using the marker-normalized maximum
liquid-surface displacement. During each experiment, the complete liquid
motion was recorded by a side-view RGB camera, and one frame was sampled
every 0.1~s for analysis. An ArUco marker attached to the container was used
for frame-to-frame registration to compensate for the relative motion between
the container and the camera. An OpenCV-based image-processing procedure was
then used to extract the liquid-surface contour on the side closer to the
camera. For each sampled frame, the maximum vertical pixel distance between
the instantaneous liquid surface and the initial equilibrium liquid surface
was computed and normalized by the detected pixel side length of the ArUco
marker in the same frame. The maximum normalized displacement over all sampled
frames was taken as the marker-normalized maximum liquid-surface displacement
for each trial. The resulting dimensionless quantities for the ten trials of each method are reported in Table~\ref{tab:liquid_surface_rw}.

\begin{figure}[pos=t]
    \centering
    \subfloat[M8]{\includegraphics[width=0.9\linewidth,height=0.28\textheight,keepaspectratio]{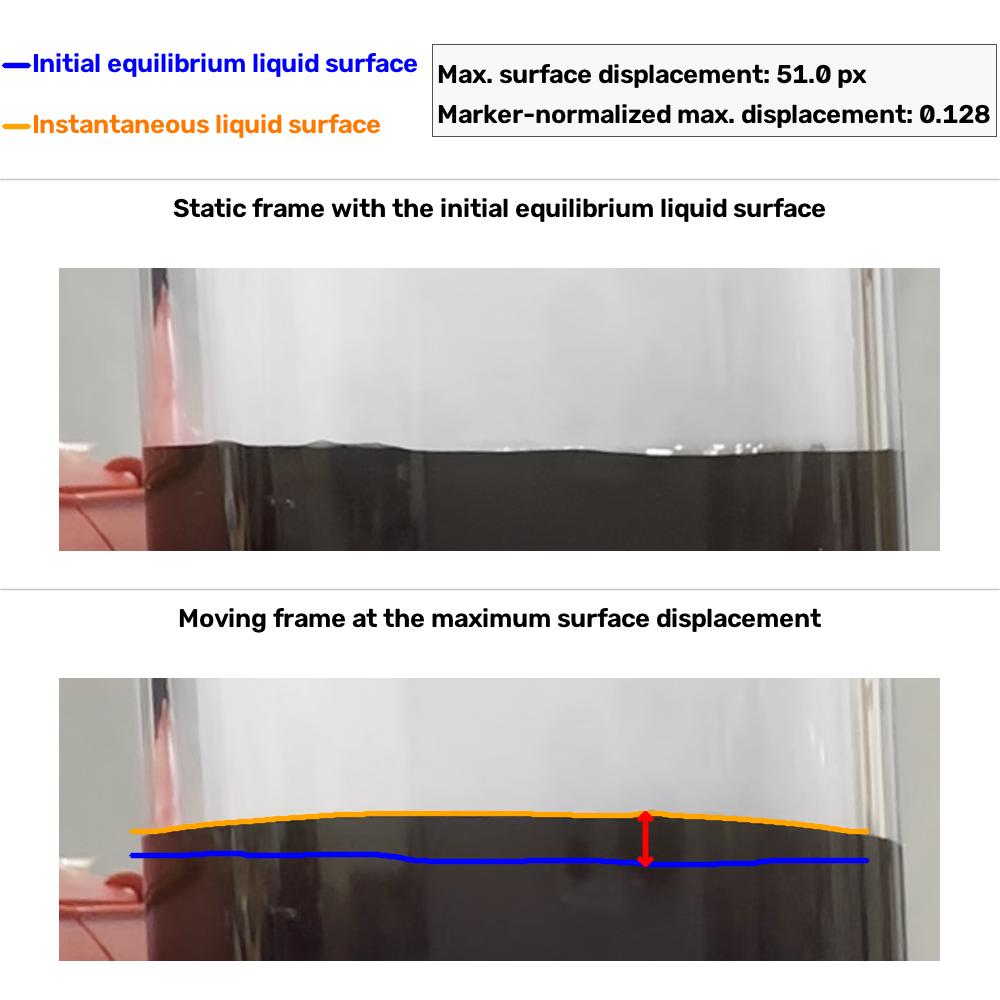}
        \label{fig:camera_m8}
    }
    \subfloat[M7]{
        \includegraphics[width=0.9\linewidth,height=0.28\textheight,keepaspectratio]{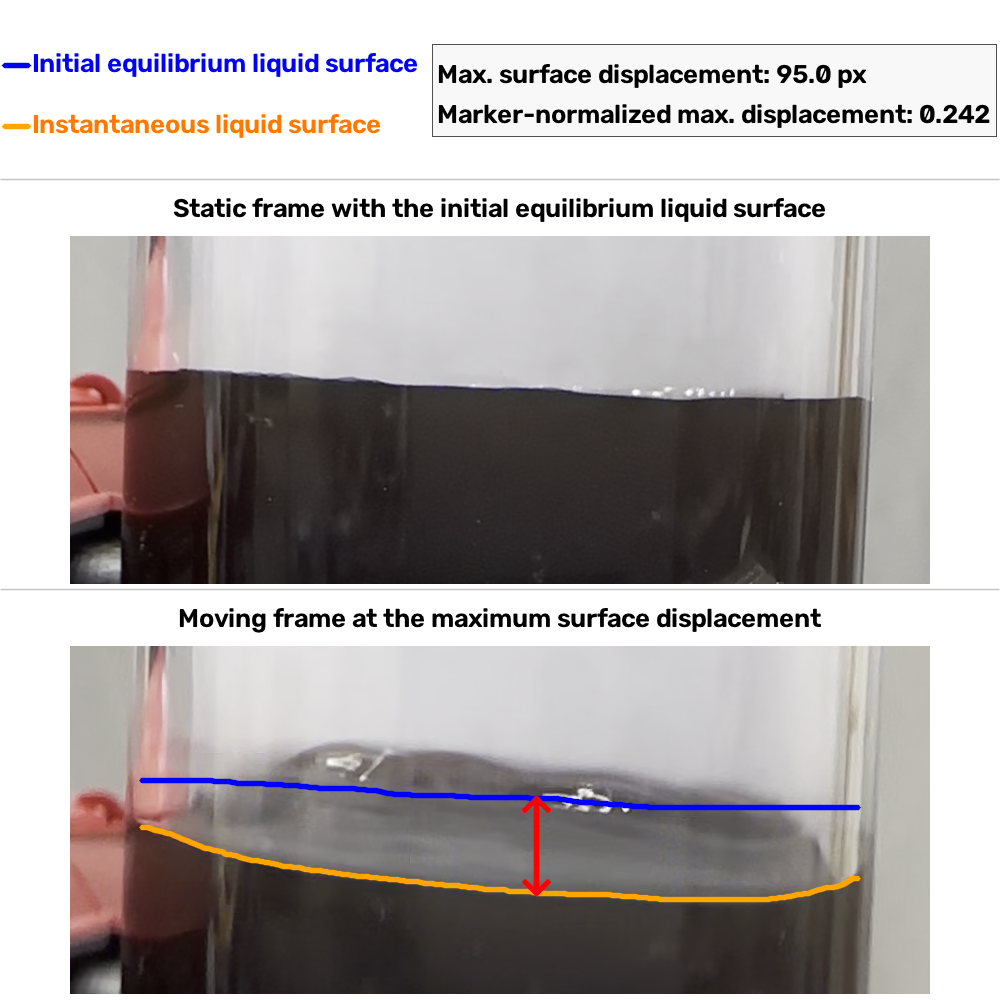}
        \label{fig:camera_m7}
    }
    \caption{Representative camera-based measurements of liquid-surface displacement for the T0 trajectory of the Between-Shelf-Level (BSL) motion: (a) M8 and (b) M7. The upper and lower images for each method show the initial equilibrium frame and the frame with the maximum detected liquid-surface displacement, respectively. The blue and orange curves denote the initial equilibrium and instantaneous liquid surfaces, and the red arrow indicates the detected maximum displacement. The corresponding pixel displacement and marker-normalized maximum liquid-surface displacement are reported at the top.}
    \label{fig:camera_liquid}
\end{figure}

In Table~\ref{tab:liquid_surface_rw}, M8 produced a lower marker-normalized maximum liquid-surface displacement than M7 in all ten paired trials. Fig.~\ref{fig:camera_liquid} presents representative camera-based measurement results for M8 and M7 on the same reference trajectory. For the BSL motion, the mean displacement decreased from \(0.150 \pm 0.054\) with M7 to \(0.115 \pm 0.021\) with M8, corresponding to a reduction of \(23.3\%\). For the IS motion, it decreased from \(0.120 \pm 0.022\) to \(0.079 \pm 0.013\), corresponding to a reduction of \(34.2\%\). These direct camera-based measurements are consistent with the lower \(E_{\mathrm{sf}}\) values obtained by M8 and provide additional quantitative evidence supporting the effectiveness of the proposed method in real-world deployment.

\subsubsection{Transport with diverse containers and liquids}

Table~\ref{tab:success_rate} summarizes the transport outcomes for the 6 container--liquid combinations, reported as successful trials over total trials. In all tested combinations, the proposed method successfully completed all 5 trials without spillage. These results indicate that the method remains effective across a representative set of container-liquid conditions and can accommodate a certain degree of variation in the manipulated objects while maintaining successful task execution.

\begin{table}[pos=t]
\centering
\caption{Comparison of \(E_{\mathrm{sf}}\) (deg$\cdot$s) between M8 and M7 in real-world experiments. BSL and IS denote between-shelf-level motion and into-shelf motion, respectively.}
\label{tab:Esf_rw}
\setlength{\tabcolsep}{3pt}
\footnotesize
\begin{tabular}{c ccccccc}
\hline
Motion & Method & T0 & T1 & T2 & T3 & T4 & Mean$\pm$Std \\
\hline
\multirow{2}{*}{BSL}
 & M8 (ours) & 3.304 & 3.033 & 2.268 & 2.678 & 3.629 & 2.982$\pm$0.531 \\
 & M7 & 6.203 & 4.594 & 5.181 & 4.292 & 5.329 & 5.120$\pm$0.739 \\
\hline
\multirow{2}{*}{IS}
 & M8(ours) & 1.829 & 2.549 & 2.922 & 1.877 & 2.134 & 2.262$\pm$0.466 \\
 & M7 & 3.494 & 4.912 & 4.533 & 3.700 & 3.211 & 3.970$\pm$0.721 \\
\hline
\end{tabular}
\end{table}

\begin{table}[pos=t]
\centering
\caption{Comparison of the marker-normalized maximum liquid-surface displacement between M8 and M7.}
\label{tab:liquid_surface_rw}
\setlength{\tabcolsep}{3pt}
\footnotesize
\begin{tabular}{c ccccccc}
\hline
Motion & Method & T0 & T1 & T2 & T3 & T4 & Mean$\pm$Std \\
\hline
\multirow{2}{*}{BSL}
 & M8 (ours) & 0.128 &0.116  &0.087 &0.102  & 0.140
 & 0.115$\pm$0.021 \\
 & M7 &0.242  &0.126  &0.104 &0.134  & 0.142
 &0.150 $\pm$ 0.054 \\
\hline
\multirow{2}{*}{IS}
 & M8 (ours) &0.068  &0.066  &0.092 &0.074  &0.093
 & 0.079$\pm$0.013\\
 & M7 &0.137  &0.115  &0.099  &0.101  & 0.150
 & 0.120 $\pm$0.022 \\
\hline
\end{tabular}
\end{table}

\begin{table}[pos=t]
\centering
\caption{Success rate of liquid transport under different container–liquid configurations.}
\label{tab:success_rate}
\begin{tabular}{lccc}
\hline
Container Type & Oolong Tea & Milk & Coca-Cola \\
\hline
Square-Rim Cup & 5/5 & 5/5 & 5/5 \\
Round-Rim Cup & 5/5 & 5/5 & 5/5 \\
\hline
\end{tabular}
\end{table}

\section{Discussion and limitations}\label{discussion}
\subsection{Limitations and extensibility of the framework}
The primary contribution of this work lies in the coordination of different modules for robotic manipulation tasks involving coupled constraints, rather than in the individual planning or control algorithms. The experimental results demonstrate the effectiveness of this coordinated, staged approach to handling coupled constraints. Nevertheless, the effectiveness of this architecture does not imply the optimality of the individual algorithms adopted in the current implementation. The framework currently assumes a known and static environment and does not accommodate moving obstacles or changes in the start or goal position during execution. Moreover, the null-space correction is a local reactive mechanism, and recovery from tracking failure is limited without online replanning. The global time-scaling strategy may also result in unnecessarily conservative motion profiles. Established planning and control techniques are adopted to evaluate the proposed architecture, but they can be replaced by alternative methods as long as the task requirements and coordination among modules are preserved. 

To accommodate environmental changes, the offline position trajectory generator in Module 1 could be replaced by B-spline-based online replanning methods \cite{9196996,WANG2025112341}, which have been successfully applied to robotic manipulators operating in dynamic environments. The B-spline representation facilitates the analytical computation of trajectory derivatives, allowing the slosh-free orientation generation in Eq.~(\ref{eq:sf orientation}) to be retained and evaluated online. Furthermore, existing incremental distance-field mapping methods \cite{10611532,10720114} could provide up-to-date environmental information for both trajectory replanning and null-space correction. Such extensions would require appropriate consideration of higher-order kinematic constraints, trajectory continuity, and manipulator feasibility. To provide more rigorous safety guarantees, a control barrier function (CBF)-based safety filter \cite{7782377} could be incorporated into the tracking controller. Importantly, such a safety filter should account for both geometric safety constraints and slosh-free orientation constraints. These extensions represent promising directions for further developing the proposed framework.

\subsection{Validity and limitations of the nominal slosh-free condition}
The term spill-free in this work refers to the task objective rather than an unconditional theoretical guarantee. The adopted nominal slosh-free condition serves as a motion-design criterion to suppress slosh-inducing excitation and thereby reduce the risk of spillage. Its effectiveness in achieving spill-free transport is demonstrated through real-world experiments.

The omission of liquid-specific physical properties in the nominal model is intended to reduce model dependence and avoid liquid-specific parameter identification. However, this simplification comes at the cost of reduced model fidelity, reflecting a trade-off between model fidelity and implementation simplicity. Liquid properties, including viscosity and surface tension, can influence actual sloshing dynamics and spillage \cite{PhysRevFluids.3.094801}, while container geometry and filling level also affect spillage conditions. These effects are not explicitly captured by the nominal model, constituting a limitation of the current approach.

Furthermore, the validity of the nominal slosh-free condition is limited by the underlying assumptions of small-amplitude sloshing and an approximately planar free surface. Although the experimental results demonstrate the effectiveness of the proposed method across the evaluated container–liquid combinations, further investigations involving a wider range of liquid properties, filling levels, and motion excitations are needed to more comprehensively characterize its applicability.

\section{Conclusion}\label{Conclusion}
This paper presented an integrated planning-and-control framework for spill-free liquid transport with open containers. The proposed method jointly considers geometric obstacle-avoidance constraints and spill-free task constraints, and combines slosh-free reference trajectory generation, learning-based initial joint configuration selection, and null-space-projection-based trajectory tracking to enable stable liquid transport in cluttered environments. Simulation and real-world experiments showed that the proposed method can successfully perform spill-free liquid transport in cluttered settings and remains effective across different container and liquid configurations.

Future work will proceed in two directions. First, we will extend the evaluation to a broader range of motion excitations and container--liquid configurations, and use direct measurements of the liquid response to further characterize the operating regime in which the adopted nominal slosh-free condition remains an adequate approximation. Second, we will explore more adaptive trajectory tracking strategies, for example by using learning-based methods to tune subtasks' weights automatically or by generating control commands directly from robot states, environment states, and reference trajectories, with the goal of further improving execution safety, stability, and adaptability.

\printcredits

\section*{Acknowledgement}
This work was supported by the National Natural Science Foundation of China under Grant 62573107, the National Science and Technology Major Project on Intelligent Manufacturing Systems and Robotics under Grant 2026ZD1609000 and the Liaoning Province Natural Science Foundation under Grant 2025-MS-051.

\section*{Declaration of competing interests}
The authors declare no competing financial or non-financial interests related to this work.

\section*{Data availability}
Data will be made available on request.

\section*{Declaration of generative AI and AI-assisted technologies in the manuscript preparation process}
During the preparation of this work, the authors used ChatGPT to improve language and readability. After using this tool, the authors reviewed and edited the content as needed and assume full responsibility for the content of the publication.

\bibliographystyle{unsrtnat}

\bibliography{references}



\end{document}